\documentclass[10pt,twocolumn,letterpaper]{article}

\usepackage{iccv}
\usepackage[table]{xcolor}
\definecolor{Gray}{gray}{0.9}

\usepackage{times}
\usepackage{epsfig}
\usepackage{graphicx}
\usepackage{amsmath}
\usepackage{amssymb}
\usepackage{tabularx}
\usepackage{booktabs, siunitx}
\usepackage{subcaption}

\usepackage[breaklinks=true,bookmarks=false]{hyperref}

\usepackage{multicol} % <===============================================

\iccvfinalcopy % *** Uncomment this line for the final submission

\def\iccvPaperID{****} % *** Enter the ICCV Paper ID here

\begin{document}

%%%%%%%%% TITLE
\title{SC-VAE: Sparse Coding-based Variational Autoencoder with Learned ISTA}

\author{Pan Xiao\qquad Peijie Qiu\qquad Sungmin Ha\qquad Abdalla Bani \qquad Shuang Zhou \qquad Aristeidis Sotiras\\
Washington University School of Medicine, Washington University in St. Louis\\
%St Louis, MO, USA\\
{\tt\small \{panxiao, peijie.qiu, sungminha, a.bani, shuangzhou, aristeidis.sotiras\}@wustl.edu}}
%\and
%Peijie Qiu\\
%Washington University in St. Louis\\
%{\tt\small aristeidis.sotiras@wustl.edu}
%\and
%Aristeidis Sotiras\\
%Washington University in St. Louis\\
%{\tt\small aristeidis.sotiras@wustl.edu}
%}

\maketitle
\ificcvfinal\thispagestyle{empty}\fi

\begin{abstract}
Learning rich data representations from unlabeled data is a key challenge towards applying deep learning algorithms in downstream tasks. Several variants of variational autoencoders (VAEs) have been proposed to learn compact data representations by encoding high-dimensional data in a lower dimensional space. Two main classes of VAEs methods may be distinguished depending on the characteristics of the meta-priors that are enforced in the representation learning step. The first class of methods derives a continuous encoding by assuming a static prior distribution in the latent space. The second class of methods learns instead a discrete latent representation using vector quantization (VQ) along with a codebook. 
However, both classes of methods suffer from certain challenges, which may lead to suboptimal image reconstruction results. The first class suffers from posterior collapse, whereas the second class suffers from codebook collapse. 
To address these challenges, we introduce a new VAE variant, termed sparse coding-based VAE with learned ISTA (SC-VAE), which integrates sparse coding within variational autoencoder framework. 
The proposed method learns sparse data representations that consist of a linear combination of a small number of predetermined orthogonal atoms. The sparse coding problem is solved using a learnable version of the iterative shrinkage thresholding algorithm (ISTA). Experiments on two image datasets demonstrate that our model achieves improved image reconstruction results compared to state-of-the-art methods. Moreover, we demonstrate that the use of learned sparse code vectors allows us to perform downstream tasks like image generation and unsupervised image segmentation through clustering image patches.
\end{abstract}
\section{Introduction}
\label{sec:intro}

A major challenge towards applying artificial intelligence to the enormous amounts of unlabeled image data gathered worldwide is the ability to learn effective visual representations of data without supervision. Various unsupervised representation learning techniques based on variational autoencoders (VAEs) \cite{kingma2013auto} have been proposed to address this challenge.
%by researchers based on the idea of variational autoencoders (VAEs) \cite{kingma2013auto}. 
The primary objective of VAEs is to learn a mapping between high-dimensional observations and a lower dimensional representation space, such that the original observations can be approximately reconstructed from the lower-dimensional representation. These lower-dimensional visual representations allow for effectively training downstream tasks, such as image generation \cite{higgins2017beta, van2017neural}, image segmentation \cite{ouyang2019data} and clustering \cite{xu2021multi, graving2020vae}. 

\begin{figure}[t!]
\centering
\includegraphics[width=7cm]{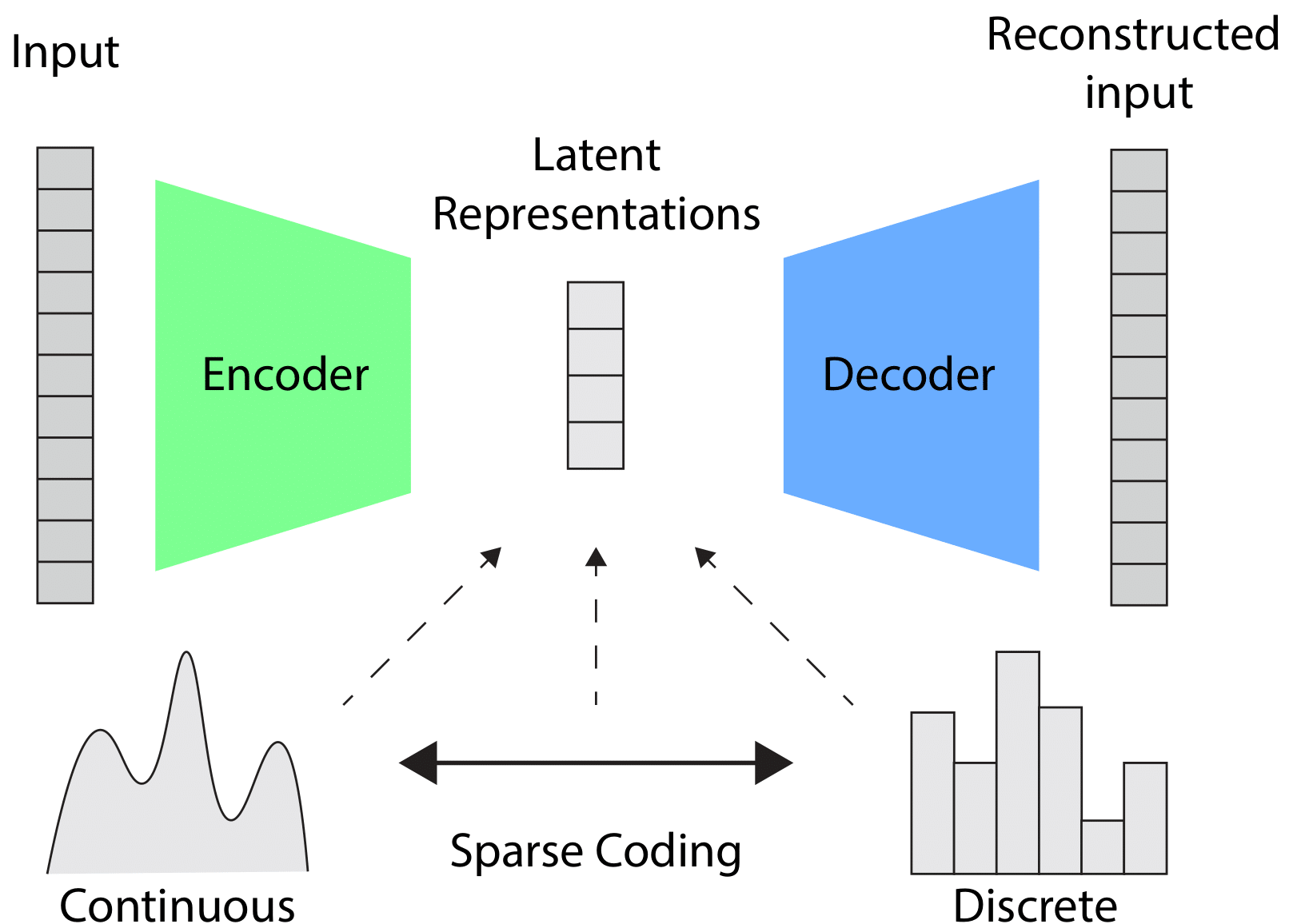}
\caption{An illustration of using sparse coding to model the latent representations of VAEs.
%The majority of VAEs assumes that the latent representations are either continuous, adhering to a static prior, or discrete via vector quantization (VQ) with a codebook.
The majority of VAEs can be categorized into two classes based on whether the latent representations are continuous, using a static prior, or discrete, utilizing vector quantization  (VQ) with a codebook.
Combining the VAE framework with sparse coding can be conceptualized as representing the middle ground between continuous and discrete VAEs.}
\label{figure:1}
\vspace{-2pt}
\end{figure}

Depending on the downstream task, different VAE variants have been proposed. These are distinguished by the assumptions they make about the world, which are encoded as meta-priors \cite{bengio2013representation}. Based on this, two main classes of methods can be identified.
The first class of methods \cite{kingma2013auto, higgins2017beta, kim2018disentangling, chen2018isolating, zhao2019infovae, kumarvariational} learns representations with continuous latent variables (hereinafter referred to as continuous VAEs). These models utilized a static prior (usually a Gaussian distribution) to regularize the latent space so that disentangled representations and diverse new data can be generated. These approaches have demonstrated good disentanglement and generation performance for simple datasets \cite{zhao2019infovae, kim2018disentangling}. However, they typically suffer from several shortcomings when applied to complex datasets. First, the static prior makes the optimization process troublesome in practice because real world datasets cannot be simply modeled by a single distribution. Second, these methods tend to ignore the latent variables if the decoder is expressive enough to model the data distribution, resulting in posterior collapse \cite{razavipreventing}. Third, while the global structure of the input is well captured by latent representations, the more intricate local structure is not, leading to bad reconstructions.

%Since the first VAEs \cite{kingma2013auto} model was proposed, most studies \cite{kingma2013auto, higgins2017beta, kim2018disentangling, chen2018isolating, zhao2019infovae, kumarvariational} focused on learning latent representations with continuous features (continuous VAEs for short). These models utilized a static prior (mostly a gaussian distribution) to regularize the latent space so that disentangled representations and diverse new data can be generated. Although demonstrating good disentanglement and generation behaviors for simple dataset, they typically suffer from several shortcomings when applied to complex datasets: 1) the static prior makes the optimization process troublesome in practice because real world datasets cannot be simply modeled by a single distribution, 2) these methods tend to ignore the latent variables if the decoder is expressive enough to model the data density, leading to posterior collapse problem \cite{razavipreventing}, 3) the global structure of the input is well captured by latent representations, but the more intricate local structure is not, causing bad reconstructions in the image space.

The second class of methods \cite{van2017neural, esser2021taming, yu2021vector, lee2022autoregressive, zheng2022movq} learns representations with discrete latent variables (hereinafter referred to as discrete VAEs).
 These methods typically utilize vector quantization (VQ) along with a codebook to learn a prior in the latent space. This approach not only circumvents issues of posterior collapse but has also been shown to achieve good image reconstruction and generation performance. Importantly, the perceptual quality of the reconstructed and generated samples  may be further improved through the use of an adversarial or perceptual loss \cite{johnson2016perceptual, larsen2016autoencoding}. However, discrete VAEs often require a large codebook to conserve the information of the encoded observations, which leads to the increase of model parameters and the codebook collapse problem \cite{dhariwal2020jukebox}. Another shortcoming of discrete VAEs \cite{van2017neural, esser2021taming, yu2021vector, lee2022autoregressive} is that they tend to generate repeated artifactual patterns in the reconstructed image because the VQ operator uses the same quantization index to embed similar image patches. Moreover, the VQ operator does not allow the gradient to pass through the codebook. This makes the optimization process more challenging, as it requires the use of techniques such as the Gumbel-Softmax trick \cite{jang2016categorical} or the straight-through estimator \cite{bengio2013estimating} to approximate the gradients.

To address aforementioned shortcomings, we introduce a new VAE variant SC-VAE, which stands for sparse coding-based VAE with learned ISTA \cite{gregor2010learning}.
Instead of using a fixed prior to learn
continuous characteristics or utilizing VQ to learn discrete variables in the latent space, we propose to model latent representations as sparse linear combinations of atoms using sparse coding (SC) \cite{rubinstein2010dictionaries}.
In the case of continuous VAEs, most of the latent factors are active at each time. In contrast, discrete VAEs require only one quantization index to be active at each time. SC-VAE adopts a middle ground, as is depicted in Figure \ref{figure:1}.
SC is anchored in the crucial realization that, despite the need for numerous variables to describe large collections of natural signals, individual instances can be effectively represented using only a small subset of these variables. 
This is in line with evidence \cite{olshausen1996emergence} that biological vision systems, such as the visual cortex in mammals, process information as sparse signals.

In the field of computer vision, SC is a widely adopted technique for efficiently reconstructing features or images \cite{bao2019convolutional, ravishankar2019image} and it has consistently demonstrated superior performance over VQ on benchmark recognition tasks \cite{coates2011importance, yang2009linear}. 
Therefore, we hypothesize that combining the VAE framework with sparse coding for latent representations reconstruction will lead to better reconstruction for the input. 
Moreover, prior works \cite{rolinek2019variational, kumar2020implicit} have indicated that incorporating orthogonality as a regularization method in VAEs is advantageous for promoting disentangled representations. 
SC-VAE offers a natural approach to acquire disentanglement by utilizing a predetermined orthogonal dictionary.
Traditional algorithms for solving the sparse coding problem, such as Iterative Shrinkage-Thresholding Algorithm (ISTA) \cite{daubechies2004iterative} and Fast Iterative Shrinkage-Thresholding Algorithm (FISTA) \cite{beck2009fast}, can be easily integrated with VAE models. However, these algorithms do not allow to back-propagate gradients, which makes it impossible to train the model in an end-to-end manner. To mitigate this issue, a learnable version of ISTA \cite{gregor2010learning} was used in this work.

Our contributions can be summarized as follows:
(1) We introduce a new VAE variant, termed SC-VAE, which seamlessly integrates the VAE framework with sparse coding. The proposed SC-VAE is trainable in an end-to-end manner and does not suffer from posterior or codebook collapse.
(2) We demonstrate that our method surpasses the performance of previous state-of-the-art approaches in image reconstruction tasks. Qualitative experiments reveal that our model exhibits superior generalization performance.
(3) By training SC-VAE with a predetermined orthogonal dictionary, we illustrate the capacity to achieve effective disentanglement and smooth interpolation in generated images through the acquired sparse code vectors.
(4) The learned sparse code vectors from SC-VAE demonstrate a capability for clustering of image patches. We highlight that this property enables us to achieve better unsupervised image segmentation performance than most state-of-the-art methods, coupled with robustness to noise.
\section{Related Work}
\label{sec:related work}
In this section, we delve into research closely related to SC-VAE. Earlier works on sparse coding-based VAEs \cite{barello2018sparse, fallah2022variational, tonolini2020variational, sadeghi2022sparsity} primarily built upon the vanilla VAE framework \cite{kingma2013auto}. These approaches achieved sparsity in latent variables by employing either specific prior distributions \cite{barello2018sparse, tonolini2020variational, sadeghi2022sparsity} or a learned threshold \cite{fallah2022variational}. For example, sparse-coding variational auto-encoder (SVAE) \cite{barello2018sparse} replaced the normal prior in the vanilla VAE with robust, sparsity-promoting priors (such as Laplace and Cauchy). However, it opted for a linear generator in the decoding process to reconstruct the input signal. This choice frequently resulted in suboptimal reconstruction outcomes due to the decoder's limited expressive capacity.
Variational sparse coding (VSC) model \cite{tonolini2020variational} focused on making the latent space interpretable by enforcing sparsity using a mixture of Spike and Slab distributions. 
%The Sparse VAE (SVAE), as introduced by Barello et al. (2018), departed from the conventional normal prior over latent variables and instead adopted heavy-tailed, sparsity-promoting priors like Laplace and Cauchy. However, a noteworthy limitation of SVAE emerged from its use of a linear generator in the decoding process to reconstruct the original input signal. This choice often resulted in suboptimal reconstruction outcomes due to the decoder's limited expressive capacity.
Variational sparse coding with learned thresholding (VSCLT) model \cite{fallah2022variational} imposed sparsity through a shifted soft-threshold function. However, the non-differentiability of this function led to numerical instability. This issue arose as the function acted as a barrier to gradient flow from the generator to the inference network when a latent feature was unused.
Sparsity-promoting dictionary model for variational autoencoders (SDM-VAE) \cite{sadeghi2022sparsity} adopted a zero-mean Gaussian prior distribution with learnable variances as a sparsity regularizer. It used a fixed dictionary with unit-norm atoms to avoid scale ambiguity.
As continuous VAEs, all these methods \cite{barello2018sparse, fallah2022variational, tonolini2020variational, sadeghi2022sparsity} faced the inherent challenge of posterior collapse. 
%To counteract this effect, a straightforward Spike variable warm-up strategy was employed in VSC \cite{tonolini2020variational}.
To counteract this effect, VSC \cite{tonolini2020variational} implemented a Spike variable warm-up strategy to prevent posterior collapse.
In contrast, the proposed SC-VAE explicitly learns sparse latent representations by combining the VAE framework with learned ISTA algorithm, avoiding the challenges of these prior works.

\section{Preliminaries}
\label{sec:preliminaries}

In this section, we first briefly introduce the optimization framework of sparse coding. Then, we describe the algorithm unrolling version of ISTA (Learnable ISTA) \cite{gregor2010learning} involved in our approach.

\subsection{Sparse Coding}
Let $X \in \mathbb{R}^n$ be an input vector,  $Z\in 
 \mathbb{R}^K$ be a sparse code vector and $\mathbf{D}\in \mathbb{R}^{n\times K}$ be a codebook matrix whose columns are atoms.
 The goal is to find an optimal way to reconstruct $X$ based on a sparse linear combination of atoms in $\mathbf{D}$.
Sparse coding typically involves minimizing the following energy function:
\begin{align}
E_{\mathbf{D}}(X,Z) = \frac{1}{2}||X-\mathbf{D}Z||_2^2 + \alpha ||Z||_1.
\label{eq:1}
\end{align}
This energy function comprises two terms. The first term is a data term that penalizes differences between the input vector and its reconstruction as quantified by the $L2$ norm. The second term  is an $L1$ norm, which acts as regularization to induce sparsity on the sparse code vector $Z$.
$\alpha$ is a coefficient
controlling the sparsity penalty. 

\begin{figure}[t!]
\centering
\includegraphics[width=8.5cm]{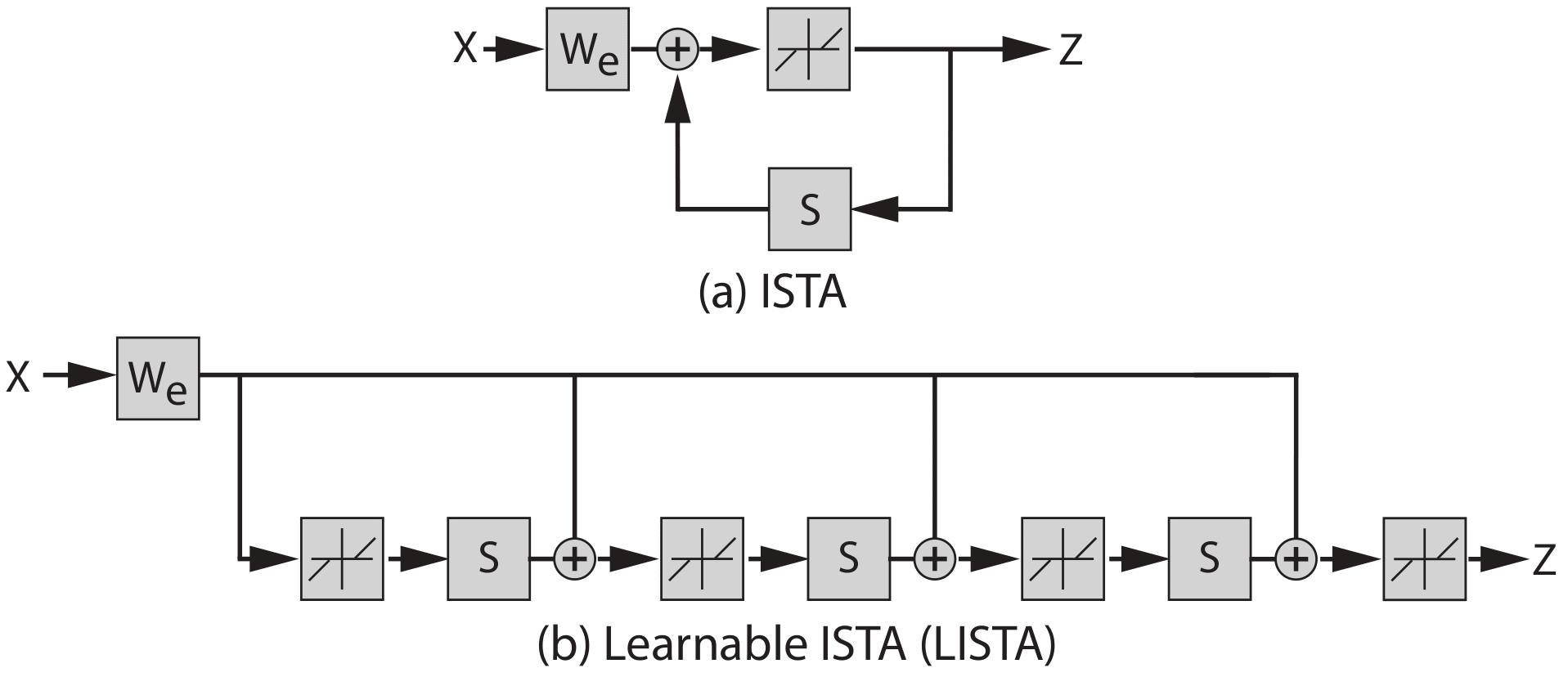}
\caption{(a) The diagram of the ISTA algorithm for
sparse coding. (b) The diagram of the Learnable ISTA, which is a time-unfolded version of the ISTA algorithm.}
\label{figure:2}
\vspace{-2pt}
\end{figure}

\begin{figure*}[htp!]
\centering
\includegraphics[width=16cm]{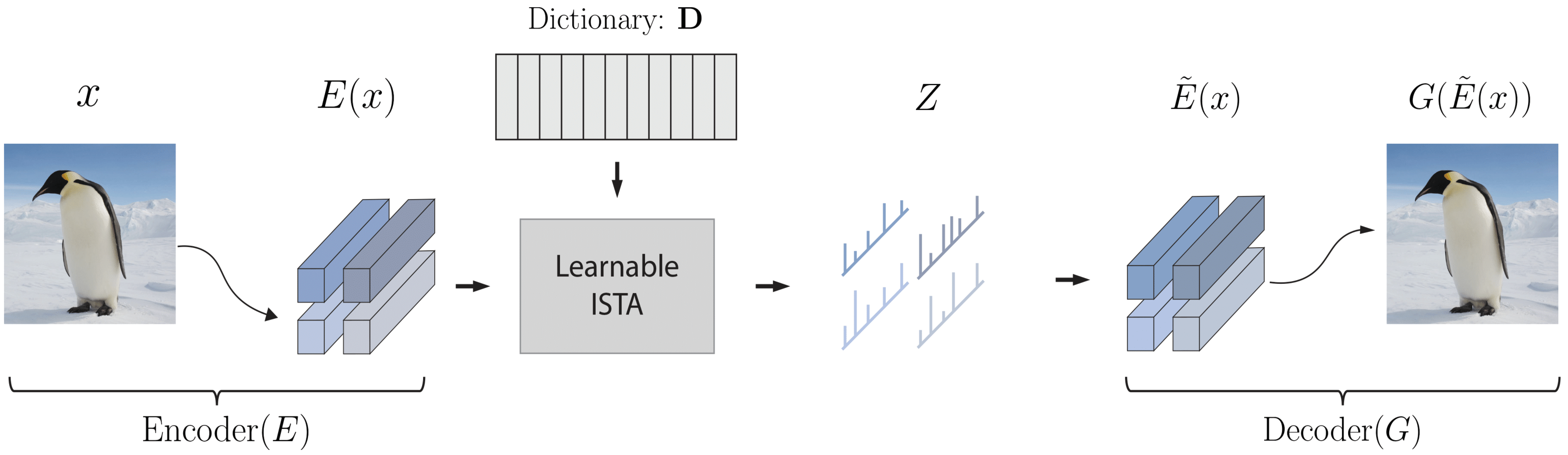}
\caption{A schematic representation of the proposed Sparse Coding-VAE with Learned ISTA (SC-VAE). SC-VAE integrates a Learnable ISTA network within VAE framework to learn sparse code vectors in the latent space for the input image. Each image can be represented as one or several sparse code vectors, depending on the number of downsampling blocks in the encoding process.}
\label{figure:3}
\vspace{-2pt}
\end{figure*}

\subsection{Learnable ISTA} \label{ISTA and LISTA}
A popular algorithm for learning sparse codes is called ISTA \cite{gregor2010learning}, which iterates the following recursive equation to convergence:
\begin{align}
Z(t+1) = h_{\theta}(W_{e}X + SZ(t))\quad Z(0) = 0.
\label{eq:2}
\end{align} 
The elements in  Eq. (\ref{eq:2}) are defined as follows:
\begin{align}
    \textup{filter matrix:} \quad &W_e = \frac{1}{L}\mathbf{D}^{\top} \nonumber\\
    \textup{mutual inhibition matrix:} \quad &S = I - \frac{1}{L}\mathbf{D}^{\top}\mathbf{D}  \nonumber\\
    \textup{shrinkage function:}\quad &[h_{\theta}(V)]_i = \textup{sign}(V_i)(|V_i|- \theta_i)_{+} \nonumber
\end{align}
Here, $L$ is a constant, which is defined as an upper bound on the largest eigenvalue of $\mathbf{D}^{\top}\mathbf{D}$.
Both the filter matrix and the mutual inhibition matrix depend on the codebook matrix $\mathbf{D}$.
The function $h_{\theta}(V)$ is a component-wise vector shrinkage function with a vector of thresholds $\theta$, where each element in $\theta$ is set to $\frac{\alpha}{L}$. 
In ISTA \cite{daubechies2004iterative}, the optimal sparse code is the fixed point of $Z(t+1) = h_{\theta}(W_{e}X + SZ(t))$. The block diagram is shown in Figure \ref{figure:2}(a).
In LISTA \cite{gregor2010learning}, $W_e$, $S$ and $\theta$ are treated as parameters of a time-unfolded recurrent neural network, where $S$ is shared over layers and the back-propagation algorithm can be performed over training samples. The number of rollout steps $s$ and the codebook matrix $\mathbf{D}$ in LISTA are predetermined. The block diagram of LISTA is shown in Figure \ref{figure:2}(b).

% \begin{algorithm}
% \caption{LISTA::fprop}\label{alg:cap}
% \begin{algorithmic}
% \Require :: fprop($X$, $Z$, $W_{e}$, $S$, $\theta$)\\
% ;; Arguments are passed by reference.\\
% ;; variables $Z(t)$, $C(t)$ and $B$ are saved for bprop.\\
% $B = W_{e}X = \frac{1}{L}\mathbf{D}^{\top}X$; \\
% $Z(0)=h_{\theta}(B)$\\
% \textbf{for} $t=1$ to $T$ \textbf{do}\\
%     \quad$C(t) = B + SZ(t-1)$\\
%     \quad$Z(t) = h_{\theta}(C(t))$\\
% \textbf{end for}\\
% $Z = Z(T)$
% \end{algorithmic}
% \end{algorithm}

%\subsection{Learnable ISTA has low rank properties} \label{LISTA}
%You need to prove that lista has low rank properties. Moreover, with the increasing of unrolling steps and parameters $\lambda$, the rank decreases and sparsity increase.

%\subsection{Why low rank properties are important}
%1. Low-rank dictionary
%learning not only enables us to provide a new data representation but also maintains feature correlation.\\
%Low-rank dictionary learning for unsupervised feature selection.\\

%\section{APPROACH}
\section{Approach}
\label{sec:approach}

The proposed SC-VAE model aims to encode an image into a series of
latent vector representations and then to utilize sparse coding to generate sparse code vectors for these representations. These sparse code vectors can be subsequently decoded back to the reconstructed image with a fixed dictionary and a decoder network. The diagram of the proposed model is shown in Figure \ref{figure:3}. We discuss the model formulation in Section \ref{Model Formulation} and the loss functions in Section \ref{Loss Functions}.

\subsection{Model Formulation} \label{Model Formulation}
 
Formally, the input of SC-VAE is an image $x \in \mathbb{R}^{H \times W \times C}$, where $H$, $W$ and $C$ denote the height, width and the number of image channels, respectively. The image
$x$ goes first through an encoder $E$ to obtain 
 latent representations $E(x) \in \mathbb{R}^{h \times w \times n}$.
Here, the values of $h$ and $w$ depend on the number of downsampling blocks $d$ in the encoder. Accordingly, these are defined as follows $h=\frac{H}{2^d}$ and $w = \frac{W}{2^d}$. $n$ denotes the number of dimensions of each latent representation $E_{ij}(x)$, where $i \in [1, h]$ and $j\in [1, w]$. $E_{ij}^{\top}(x) \in \mathbb{R}^{n \times 1}$ is then given as an input to a Learnable ISTA network. The Learnable ISTA produces the sparse code vector $Z_{ij}\in \mathbb{R}^{1\times K}$ for each $E_{ij}(x)$ by using the learnable parameters $W_e$, $S$ and $\theta$. 
Here, $K$ denotes the number of atoms in the  predetermined orthogonal dictionary $\mathbf{D}$. Each reconstructed latent representation $\tilde{E}_{ij}(x)$ can be calculated by the multiplication of $Z_{ij}$ and $\mathbf{D}^{\top}$, which is then used to reconstruct the original image by going through a decoder neural network $G$.
We denote the output of SC-VAE as $G(\tilde{E}(x))$.
%The design of our convolutional encoder $E$ and decoder $G$ follows the architecture in \cite{esser2021taming}. 

\subsection{Loss Functions} \label{Loss Functions}
We need
to define loss functions at two levels: the image level and the latent representation level. The loss in the image level should encourage our model to provide a good reconstruction for the input image. The loss in the latent space should  allow us not only to obtain good latent representation reconstruction, but also to learn sparse codes. 
%which can reconstruct the latent representations well.

\noindent\textbf{Image reconstruction.} The most common image reconstruction term utilized in VAE models is the $L2$ loss. The $L2$ loss is defined as
\begin{align} \label{eq:4}
    \mathcal{L}_{rec}  &= ||G(\tilde{E}(x))-x||_2^2.
\end{align}

\noindent
\textbf{Latent representation reconstruction.} 
We aim to learn how to reconstruct each latent representation $E_{ij}(x)$ based on a linear combination of atoms in the fixed orthogonal dictionary $\mathbf{D}$.
Accordingly, the loss function for the latent representation reconstruction is given by:
\begin{align} \label{eq:4}
    \mathcal{L}_{latent}  =\sum_{ij}(||E_{ij}^{\top}(x) -\mathbf{D}Z_{ij}^{\top}||_2^2  + \alpha||Z_{ij}^{\top}||_1).
\end{align}
Similar to Eq. (\ref{eq:1}), this loss consists of two terms. The first term is a $L2$ norm to penalize differences between the latent representations of input images and their latent representation reconstructions. The second term imposes sparsity to each latent sparse code vector $Z_{ij}$. $\alpha$ controls the sparseness of the learned sparse code vectors $Z$.

\noindent\textbf{Total loss.} An intuitive way to build the total loss function would be to simply add $\mathcal{L}_{rec}$ and $\mathcal{L}_{latent}$.
However, this loss function will not allow us to learn a good image reconstruction due to the summation term in $\mathcal{L}_{latent}$.
This is because each input image corresponds to $h\times w$ latent representations. As a consequence, the model will focus on learning good sparse code vectors for these latent representations and pay less attention on  adequately optimizing $\mathcal{L}_{rec}$. 
To account for this factor, we introduce coefficients $\frac{1}{hw}$ to each of the latent representation $E_{ij}(x)$, which allows for appropriately balancing the two terms.
Thus, the total loss for our model is the following:
\begin{align} \label{eq:6}
    \mathcal{L}_{SC-VAE}(E,G,W_e, S, \theta)  =\mathcal{L}_{rec} +  \frac{1}{hw}\mathcal{L}_{latent}.
\end{align}
\section{Experiments}
\label{sec:experiments}

In this section, we first provide summaries of the datasets used to train the SC-VAE model
and elaborate on the implementation details. Then, we demonstrate the effectiveness of SC-VAE under different problem settings,  including 1) image reconstruction, 2) image generation through manipulating and interpolating learned sparse code vectors, 3) image patches clustering and 4) unsupervised image segmentation associated with its noise robustness analysis.
Moreover, an ablation study for the influence of the number of rollout steps (s) in LISTA is provided.

\noindent\textbf{Dataset.} For our experiments, we used the Flickr-Faces-HQ (FFHQ) \cite{karras2019style} and ImageNet \cite{deng2009imagenet} datasets. 
%The FFHQ dataset is a collection of high-quality images featuring human faces, which exhibits a significant amount of diversity in terms of the subjects' age, ethnicity, and the backgrounds depicted in the images. 
The FFHQ dataset comprises $70,000$ images with a training set of $60,000$ images and a validation set of $10,000$ images. ImageNet is a widely used benchmark dataset for visual recognition tasks. It consists of $1.2$ million training images and $50,000$ validation images, with each image associated with one of $1,000$ distinct categories. %The images in the dataset are high-resolution and have a wide variety of objects, scenes, and backgrounds, making it a challenging and diverse dataset for VAEs to model. 
The images in both datasets are high-resolution and diverse, making them challenging for VAEs to model. All images were resized to $256\times 256$ pixels.

\noindent\textbf{Implementation Details.} We adopted the encoder and decoder architecture of the VQGAN \cite{esser2021taming}.
We trained SC-VAE models on the FFHQ and ImageNet datasets with four different number of downsampling blocks.
The downsampling blocks $d$ of the encoder were set to $3$, $4$, $6$ and $8$, resulting in $32\times 32$, $16\times 16$, $4\times 4$ and $1\times 1$ sparse code vectors in the latent space for an input image with a resolution of $256\times 256\times 3$.
Moreover, the number of atoms in the dictionary, the number of rollout steps in LISTA and the dimension of latent vector representations were set to $512$, $16$ and $256$, respectively. 
%The number of atoms in the dictionary and the unfolded blocks in LISTA were set to $512$ and $5$ respectively. 
%The overall structure of the attention network $F$ was given by the following expression, in which $[a\times b ]$ symbolizes a multiplication by a matrix of that size: $E_{ij}(x)-\mathbf{D}Z_{ij}\rightarrow [256\times 64]\rightarrow \textup{Sigmoid} \rightarrow [64\times 1] \rightarrow \textup{Softmax} \rightarrow \alpha_{ij}$. The overcomplete
Discrete Cosine Transform (DCT) orthogonal matrix was used to generate the dictionary.
The sparsity coefficient $\alpha$ was initialized to $1$ and then updated during training.
To optimize our SC-VAE model, the Adam \cite{kingma2014adam} optimizer was used with a learning rate of $10^{-4}$. The training  ran for a total of $10$ epochs and the batch size for each iteration was set to $16$. We kept the models that achieved the lowest total loss for further evaluation. 
More detailed information about the architecture of SC-VAE along with a visualization of dictionary atoms can be found in the supplementary material.

\subsection{Image Reconstruction}
\noindent\textbf{Baselines and Evaluation Metrics.} Two sparse coding-based VAEs (VSC \cite{tonolini2020variational}  and VSCLT \cite{fallah2022variational} ), three continuous VAE (Vanilla VAE \cite{kingma2013auto}, $\beta$-VAE \cite{higgins2017beta}, and Info-VAE \cite{zhao2019infovae}) and four discrete VAE models (VQGAN \cite{esser2021taming}, ViT-VQGAN \cite{yu2021vector}, RQ-VAE \cite{lee2022autoregressive} and Mo-VQGAN \cite{zheng2022movq}) were selected as our baselines. 
%The model architecture used is the same as in the original paper.
We used the same model architectures as the ones described in the respective papers.
%Four most common evaluation metrics including PSNR, SSIM, LPIPS \cite{zhang2018unreasonable} and rFID \cite{heusel2017gans} were used to 
We evaluated the quality between reconstructed images and original images using four most common evaluation metrics (i.e., Peak Signal-to-Noise Ratio (PSNR), Structural Similarity Index Measure (SSIM), Learned Perceptual Image Patch Similarity (LPIPS) \cite{zhang2018unreasonable}, and reconstructed Fréchet Inception Distance (rFID) \cite{heusel2017gans}). PSNR measures the amount of noise introduced by the reconstruction process. SSIM quantifies the similarity between two images by taking into account not only pixel values, but also the structural and textural information in the images. LPIPS and rFID use a pre-trained deep neural network to measure the perceptual distance and distribution distance between two images, respectively.

%Without sparsity presented
\begin{table}[!htbp]
\centering
\caption{Quantitative reconstruction results on the validation sets of FFHQ \cite{karras2019style} ($10,000$ images) and ImageNet \cite{deng2009imagenet} ($50,000$ images). 
%Comparison of PSNR, SSIM, LPIPS and rFIDs between  validation images and their reconstructed images according to the cookbook size ($K$) and the shape of latent codes ($\mathcal{C})$.
$\mathcal{C}$ and $K$ denote the shape of latent codes and the cookbook size, respectively.
$\dag$, $\ddag$, $\curlyvee$ and $\curlywedge$ are used to distinguish the same method with different $\mathcal{C}$. The top two results across different metrics are highlighted in bold and red, respectively.} 
\resizebox{1.0\linewidth}{!}{%
\begin{tabular}{c|c|c|c|c|c|c|c}
  \toprule
  Model & Dataset & $\mathcal{C}$ & $K$  & PSNR $\uparrow$ & SSIM $\uparrow$ & LPIPS $\downarrow$ & rFID $\downarrow$\\
  \midrule
  VSC \cite{tonolini2020variational} &  & $256$ & -  & $17.62$ & $0.4458$ & $0.6148$ & $450.98$\\
  VSCLT \cite{fallah2022variational}  &  & $256$ & -  & $12.97$ & $0.3034$ & $0.4426$ & $249.54$\\
  Vanilla VAE \cite{kingma2013auto} &  & $256$ & -  & $18.00$ & $0.4960$ & $0.6568$ & $178.17$ \\
  $\beta$-VAE \cite{higgins2017beta} &  & $256$ & -  & $16.63$ & $0.4763$ & $0.6878$ & $186.87$ \\
  Info-VAE \cite{zhao2019infovae} &  & $256$ & -  & $15.85$ & $0.4560$ & $0.6990$ & $232.37$ \\
  VQGAN \cite{esser2021taming} & & $16 \times 16 \times 1$ & $1024$  & $22.24$ & $0.6641$ & $0.1175$ & $4.42$\\
  ViT-VQGAN \cite{yu2021vector} & FFHQ & $32 \times 32 \times 1$ & $8192$  & - & - & - & $\textcolor{red}{3.13}$\\
  RQ-VAE$^{\dag}$ \cite{lee2022autoregressive} & & $8 \times 8 \times 4$ & $2048$  & $22.99$ & $0.6700$ & $0.1302$ & $7.04$ \\
  RQ-VAE$^{\ddag}$ \cite{lee2022autoregressive} &  & $16 \times 16 \times 4$ & $2048$   & $24.53$ & $0.7602$ & $0.0895$ & $3.88$ \\
  Mo-VQGAN \cite{zheng2022movq} &  & $16 \times 16 \times 4$ & $1024$  & $26.72$ & $0.8212$ & $\textcolor{red}{0.0585}$ & $\textbf{2.26}$ \\
  \rowcolor{Gray}
  SC-VAE$^{\dagger}$ & & $1 \times 1\times 1$ & $512$  & $17.25$ & $0.4946$ & $0.7177$ & $195.30$\\
  \rowcolor{Gray}
  SC-VAE$^{\ddag}$ & & $4 \times 4\times 1$ & $512$  & $22.29$ & $0.6110$ & $0.4907$ & $115.91$\\
  \rowcolor{Gray}
  SC-VAE$^{\curlyvee}$ & & $16 \times 16\times 1$ & $512$ & $\textcolor{red}{29.70}$ & $\textcolor{red}{0.8347}$ & $0.1956$ & $41.56$\\
  \rowcolor{Gray}
  SC-VAE$^{\curlywedge}$ & & $32 \times 32 \times 1$ & $512$ &  $\textbf{34.92}$ & $\textbf{0.9497}$ & $\textbf{0.0080}$ & $4.21$\\
  \midrule
  VSC \cite{tonolini2020variational} &  & $256$ & - & $17.76$ & $0.5534$ & $0.5999$ & $454.75$\\
  VSCLT \cite{fallah2022variational} &  & $256$ & - & $12.70$ & $0.3016$ & $0.7578$ & $336.72$\\
  Vanilla VAE \cite{kingma2013auto} &  & $256$ & - & $17.88$ & $0.4441$ & $0.6957$ & $163.00$ \\
  $\beta$-VAE \cite{higgins2017beta} & & $256$ & - & $16.10$ & $0.4165$ & $0.7142$ & $224.03$ \\
  Info-VAE \cite{zhao2019infovae} &  & $256$ & - & $17.20$ & $0.4259$ & $0.7097$ & $178.94$ \\
  VQGAN$^{\dag}$ \cite{esser2021taming} &  & $16 \times 16 \times 1$ & $1024$ & $19.47$ & $0.5214$ & $0.1950$ & $6.25$\\
  VQGAN$^{\ddag}$ \cite{esser2021taming} &  & $16 \times 16\times 1$ & $16384$ & $19.93$ & $0.5424$ & $0.1766$ & $3.64$\\
  %VQGAN$^{\S}$ &  & $32 \times 32 \times 1$ & $8192$ & $1.21$ & $23.55$ & $0.68$ \\
  ViT-VQGAN \cite{yu2021vector} & ImageNet & $8 \times 8 \times 4$ & $2048$  & - & - & - & $1.28$\\
  RQ-VAE \cite{lee2022autoregressive} &  & $8 \times 8 \times 4$ & $2048$ & $20.76$ & $0.5416$ & $0.1381$ & $4.73$\\
  Mo-VQGAN \cite{zheng2022movq} & & $16 \times 16 \times 4$ & $1024$  & $22.42$ & $0.6731$ & $\textcolor{red}{0.1132}$ & $\textcolor{red}{1.12}$\\
  \rowcolor{Gray}
  SC-VAE$^{\dag}$ & & $1 \times 1\times 1$ & $512$  & $16.86$ & $0.4306$ & $0.8163$ & $185.68$\\
  \rowcolor{Gray}
  SC-VAE$^{\ddag}$ & & $4 \times 4\times 1$ & $512$ & $23.04$ & $0.5906$ & $0.5081$ & $114.08$\\
  \rowcolor{Gray}
  SC-VAE$^{\curlyvee}$ & & $16 \times 16\times 1$ & $512$  & $\textcolor{red}{30.74}$ & $\textcolor{red}{0.8594}$ & $0.1447$& $23.23$\\
  \rowcolor{Gray}
  SC-VAE$^{\curlywedge}$ & & $32 \times 32 \times 1$ & $512$  & $\textbf{38.40}$ & $\textbf{0.9688}$ & $\textbf{0.0070}$& $\textbf{0.71}$\\
  \bottomrule
\end{tabular}}

\label{Table_rec}
\end{table}
\begin{figure}[tbp]
\centering
\includegraphics[width=8cm]{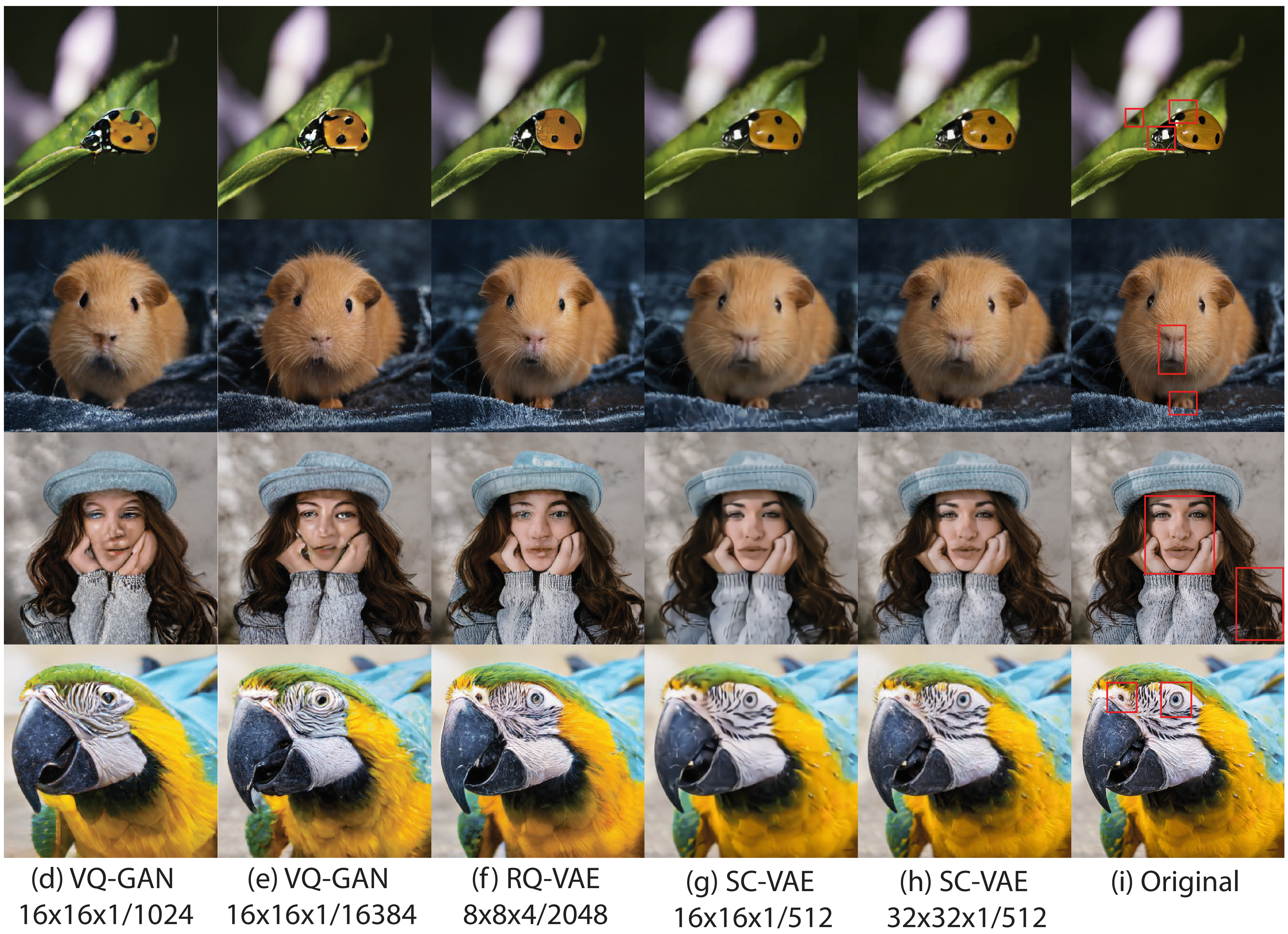}
\caption{Image reconstructions from different models trained on ImageNet dataset. Original images in the top two rows are from the validation set of ImageNet dataset. Two external images are shown in the last two rows to demonstrate the generalizability of different methods. The numbers denote the shape of latent codes and learned codebook (dictionary) size, respectively. SC-VAE  achieved improved image reconstruction compared to the baselines. Zoom in to see the details of the red square area.}
\label{figure:rec}
\vspace{-2pt}
\end{figure}
Quantitative experimental results comparing the image reconstruction performance of SC-VAE with the baseline methods are listed in Table \ref{Table_rec}. We report the performance of our model with four different downsampling blocks ($d=3, 4, 6, 8$). The top two results across different metrics are highlighted in bold and red, respectively. As can be seen in the Table \ref{Table_rec}, SC-VAE significantly improved the image quality compared to other methods in terms of PSNR, SSIM and LPIPS scores in the FFHQ dataset and in terms of all scores in the ImageNet dataset when the shape of latent codes was set to $32\times32\times 1$.
%Our method outperformed other methods with a large margin in terms of PSNR and SSIM scores in the FFHQ dataset and in terms of PSNR, SSIM and LPIPS scores in the ImageNet dataset. 
Even after downsampling the original image to a shape of $16\times16\times1$, our approach significantly outperformed other methods in terms of PSNR and SSIM scores on both datasets. With increasing number of downsampling blocks, our model struggled with image reconstruction.

Among baseline methods, sparse coding-based VAEs (i.e. VSC \cite{tonolini2020variational} and VSCLT \cite{fallah2022variational}) and continuous VAEs (i.e. Vanilla VAE \cite{kingma2013auto}, $\beta$-VAE \cite{higgins2017beta}, and Info-VAE \cite{zhao2019infovae}) produced poor reconstructions of  the original images. This is because both FFHQ and ImageNet datasets exhibit a significant amount of diversity in terms of objects and background, which is challenging to model with a static prior distribution in the latent space. Discrete VAEs (i.e. VQGAN \cite{esser2021taming}, ViT-VQGAN \cite{yu2021vector}, RQ-VAE \cite{lee2022autoregressive} and Mo-VQGAN \cite{zheng2022movq}) demonstrated better performance compared to continuous VAEs. However, they produced lower PSNR and SSIM scores compared to our model. That might be because the adversarial training and perceptual loss of discrete VAEs force the model to generate more visually appealing and realistic outputs without paying sufficient attention to structural information. Furthermore, 
these models used a much larger codebook size than ours. 
%the codebooks of these models are considerably larger in size compared to ours.
The codebook collapse problem caused by the larger codebook might make it difficult for them to reconstruct detailed and accurate information.

Four images and paired reconstructed results from different models  were visualized in Figure \ref{figure:rec}.
%The first two rows display images that have been reconstructed by various models, as well as original images from ImageNet's validation set.
%Four reconstructed images across different models were visualized in Figure \ref{figure:4}. 
The models used to reconstruct these images are trained on the training set of ImageNet dataset. 
%The results from VIT-VQGAN and Mo-VQGAN were not given due to the unavailability of official code and pre-trained models.
The absence of official code and pre-trained models prevents us from providing results for VIT-VQGAN and Mo-VQGAN. We present visualizations from unofficial implementation of VIT-VQGAN in the supplementary materials for reference.
As for sparse coding and continuous VAEs, we refrained from presenting their visualizations due to their subpar reconstruction capabilities.
As is shown in the Figure \ref{figure:rec}, 
%the reconstructed results from Vanilla VAE, $\beta$-VAE, and Info-VAE were blurry. 
VQ-GAN, RQ-VAE and our models accomplished visually appealing outcomes. 
%Top two rows show the reconstructed images of different models and original images from the validation set of ImageNet.
The original images in the first two rows were randomly selected from the validation set of ImageNet. VQGAN tended to generate repeated artifacts on similar semantic patches, such as the leaves and noses. While RQ-VAE enhanced the visual aspect, it struggled to accurately reconstruct intricate details and complex patterns. In contrast to these models, SC-VAE generated much more authentic-looking details without suffering from any of the aforementioned shortcomings.
The original images in the last two rows are part of the external images of ImageNet dataset. They were used to test the generalizability of different models. SC-VAE was able to accurately identify and reconstruct patterns that have not previously been encountered. However, the reconstructed images from VQGAN and RQVAE did not preserve the detailed information and distorted the original images.

\subsection{Image Generation}
\label{image_generation}
\begin{figure}[tbp]
\centering
\includegraphics[width=8.2cm]{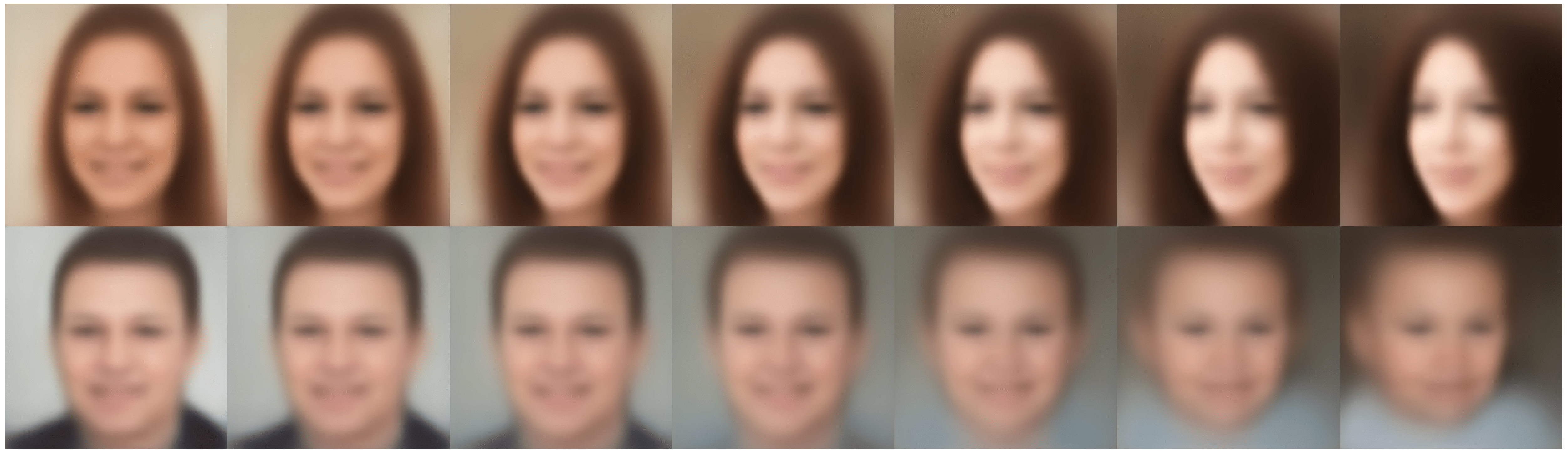}
\caption{Interpolation between the sparse code vectors of two samples from the SC-VAE$^{\dag}$ model trained on FFHQ.}
\label{figure:interpolation}
\vspace{-2pt}
\end{figure}

%In this section, we evaluate SC-VAE on image generation in terms of both manipulating and interpolating sparse code vectors.\\
%Nevertheless, we will illustrate in Section \ref{image_generation} that SC-VAE exhibits disentanglement behavior at the expense of its reconstruction capability.\\
In this section, we explore whether utilizing a pre-defined orthogonal DCT matrix in the SC-VAE model enables the generation of images by manipulating and interpolating sparse code vectors.\\
\noindent\textbf{Manipulating sparse code vectors.}
\begin{figure}[tbp]
\centering
\includegraphics[width=7.6cm]{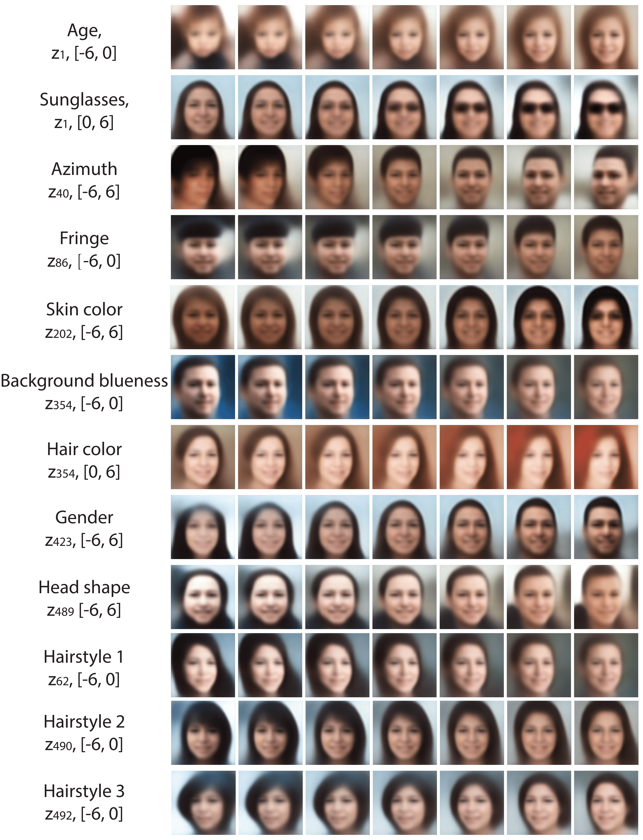}
\caption{Manipulating sparse code vectors on FFHQ. Each row represents a different seed image used to infer
the latent sparse code vector in the SC-VAE$^\dag$ model.
The disentangled attributes associated with the $i$-th component of a sparse code vector $z$ and a traversal range are shown in the first column.}
\label{figure:manipulation}
\vspace{-2pt}
\end{figure}
%It is not possible to quantify feature disentanglement in natural data, as the source features are not known. However, similarly to \cite{kim2018disentangling}, we can qualitatively examine the effect of changing single latent variables on generated samples. 
Quantifying feature disentanglement in natural data is challenging since the source features are typically unknown. Nevertheless, as in the approach of \cite{kim2018disentangling}, we can assess the impact of altering individual components of a sparse code vector on the generated samples qualitatively.
%To this end, we train a VSC model with 100, 000 examples from the CelebA data set, encode examples from a test set, alter individually exploited dimensions in the latent space and finally generate samples from these altered latent vectors. 
To this end, we used the SC-VAE$^\dag$ model with a downsampling block of $8$ trained on the FFHQ dataset. 
%encode examples from a validation set, alter individually exploited dimensions in the sparse code vectors and finally generate samples from these altered latent sparse vectors. 
We selected examples from the validation set, manipulated individual dimensions within the sparse code vectors, and then produced samples based on these modified latent sparse vectors.
%We find that several of the dimensions exploited by the VSC model control interpretable aspects in the generated data, as shown in the examples of Figure \ref{figure:manipulation}.
We found that $10$ components of the dimensions exploited by the SC-VAE$^\dag$ model controlled interpretable aspects in the generated data, as is shown in Figure \ref{figure:manipulation}.
%The $1$-st and $354$-th dimensions contains two interpretable attributes, as is shown in the examples of Figure \ref{figure:manipulation}. 
%To the best of our knowledge, we are the first one to report the disentanglement behaviour of a deterministic VAE.
\\
\noindent\textbf{Interpolating sparse code vectors.}
Figure \ref{figure:interpolation} shows
smooth interpolation between the latent sparse code vectors of two images generated by SC-VAE$^\dag$.\\
Additional manipulation and interpolation results can be found in the supplementary material.

\subsection{Image Patches Clustering}
\begin{figure}[t!]
\centering
\includegraphics[width=8.2cm]{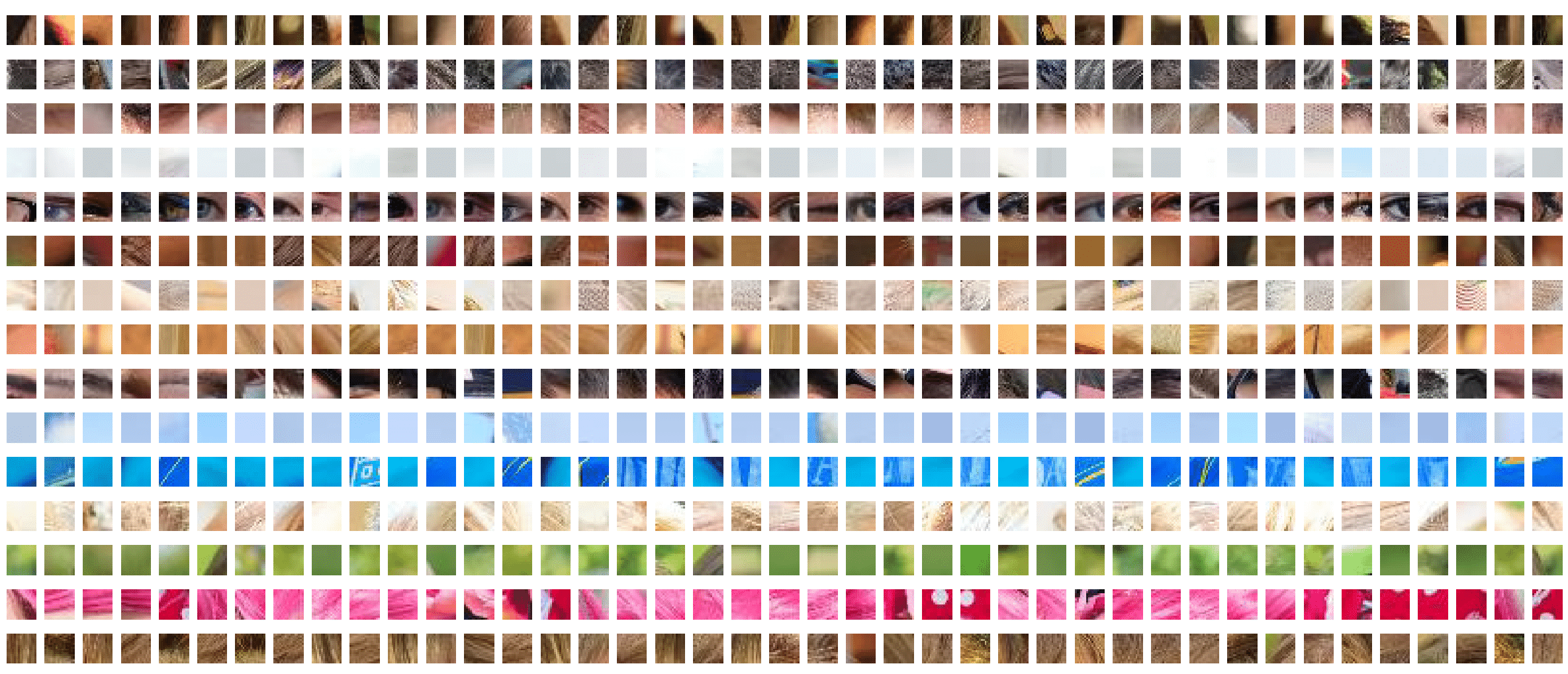}
\caption{$15$ randomly selected image patch clusters generated by clustering the learned sparse code vectors of SC-VAE$^\curlyvee$ using the K-means algorithm. Each row represents one cluster. Image patches with similar patterns were grouped together.}
\label{figure:image_patches_clustering}
\vspace{-2pt}
\end{figure}

The latent sparse code vectors learned by SC-VAE can be thought of as compressed representations of the input image. To better interpret the learned sparse code vectors, we aligned each of them with a corresponding patch of the input image. Image patch clustering was then performed based on these sparse code vectors.
We examined one pre-trained SC-VAE$^\curlyvee$ model on FFHQ dataset with a downsampling block of $d=4$. $1,000$ images were randomly selected from the validation set, resulting in $256,000$ pairs of image patches with a resolution of $16\times 16\times 3$ and sparse code vectors with a dimension of $512$. The sparse code vectors were clustered into $1,000$ groups using the K-means clustering algorithm. We randomly selected $15$ groups and $40$ patches from each group to visualize the clustering results. As can be seen in Figure \ref{figure:image_patches_clustering}, image patches with similar patterns were grouped together. More clustering results on the validation set of FFHQ and ImageNet datasets can be found in the supplementary material.

\begin{figure}[tbp]
\centering
\includegraphics[width=8cm]{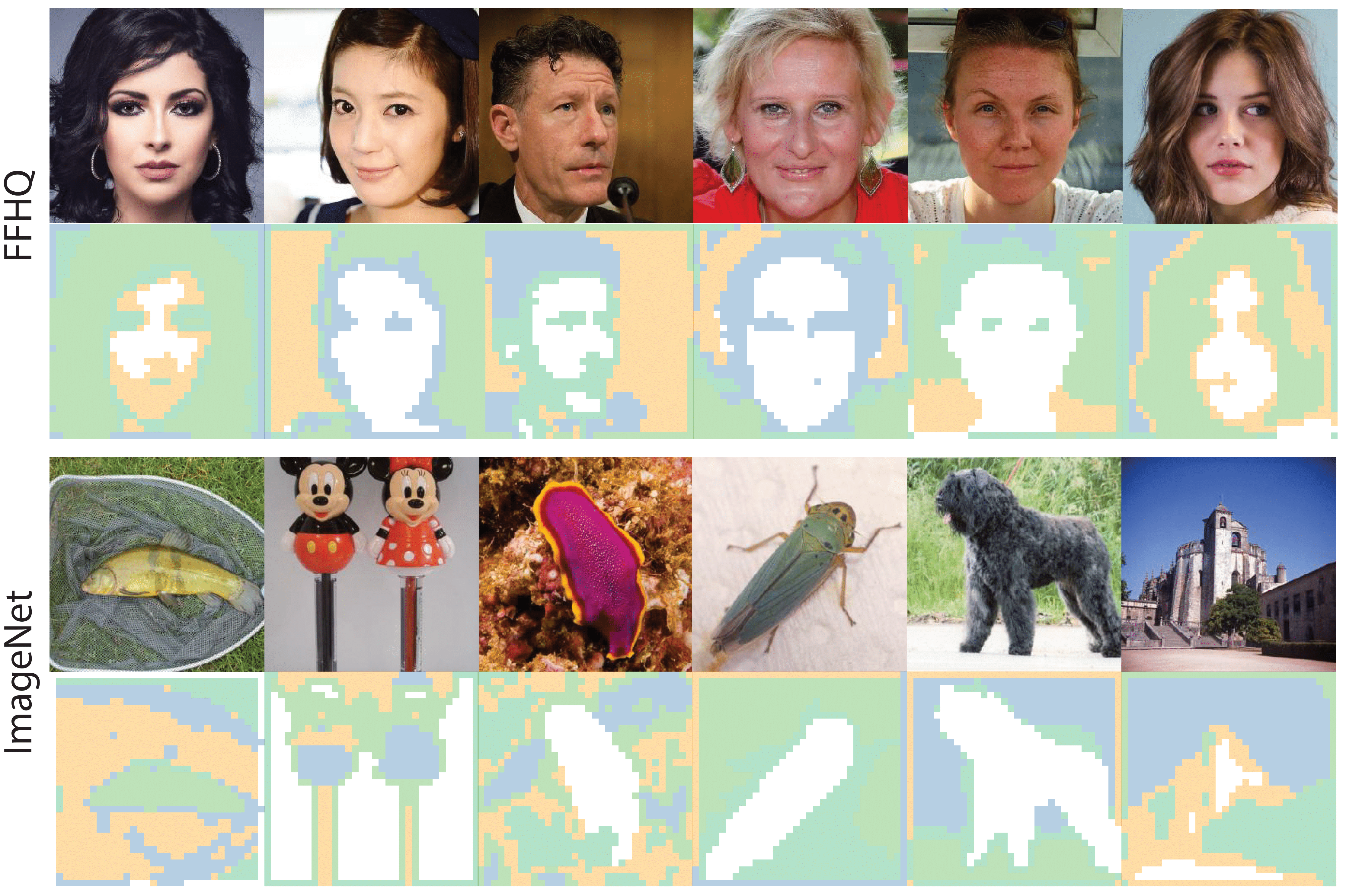}
\caption{Unsupervised image segmentation results. \textbf{Top}: images from the validation set of FFHQ. \textbf{Bottom}: images from the validation set of ImageNet.}
\label{figure:UIS_quali}
\vspace{-2pt}
\end{figure}

\subsection{Unsupervised Image Segmentation}
This section investigates whether the sparse code vectors estimated by our models enable us to conduct image segmentation tasks without supervision.
\\
\textbf{Qualitative analysis.} We utilized two SC-VAE$^\curlywedge$ models that were pre-trained on the training set of the FFHQ and ImageNet datasets, respectively. These models had a downsampling block of $d = 3$. The images were transformed into sparse code vectors with a size of $32 \times 32 \times 1$ using the encoder and LISTA network of
the SC-VAE$^\curlywedge$. Afterwards, the K-means algorithm was applied to cluster these sparse code vectors into $5$ categories, generating a $32 \times 32$ mask with $5$ different classes for each image. The segmentation results were visualized in Figure \ref{figure:UIS_quali}. The faces and objects in the images were successfully detected and segmented by simply grouping the patch-level sparse codes using the K-means algorithm.\\
\noindent
\textbf{Quantitative comparisons to prior work.} 
We evaluated the SC-VAE$^\curlywedge$ model pre-trained on ImageNet dataset on Flowers \cite{nilsback2008automated}, Caltech-UCSD Birds-200-2011 (CUB) \cite{WahCUB_200_2011} and  International Skin Imaging Collaboration 2016 (ISIC-2016) \cite{gutman2016skin} datasets. 
%The SC-VAE model was pre-trained on ImageNet dataset with a downsampling block of $d = 3$.
%We compared its performance to prior works (see Table \ref{tbl_un}).
%We evaluated SC-VAE on Caltech-UCSD Birds-200-2011 (CUB) \cite{he2022ganseg} and Flowers \cite{he2022ganseg} datasets and compared its performance to prior works (see Table \ref{tbl_un}). 
We followed \cite{savarese2021information} to generate the test set and the ground-truth masks of Flowers \cite{nilsback2008automated} and CUB \cite{WahCUB_200_2011}. 
%The unsupervised segmentation results were obtained by applying a spectral clustering algorithm on sparse code vectors. 
%To obtain the unsupervised image segmentation results, we first used a spectral clustering algorithm to cluster the sparse code vectors into $2$ or $3$ classes per image. Then the boundary connectivity \cite{zhu2014saliency} was used to decide if each cluster belongs to the foreground or background.
To achieve unsupervised image segmentation results, we initially applied a spectral clustering algorithm to group the sparse code vectors into 2 or 3 classes per image. Subsequently, we utilized boundary connectivity information \cite{zhu2014saliency} to determine whether each class corresponds to the foreground or background.
We compared the performance of our method to prior works (see Table \ref{tbl_un}). 
Note that this is not a fair comparison.
Most of the methods we compare to, excluding GrabCut \cite{rother2004grabcut} and IEM \cite{savarese2021information}, require a training set with a distribution identical to that of the test set for model training. In contrast, our pre-trained SC-VAE$^\curlywedge$ model can segment an image without the need for fine-tuning on Flower \cite{nilsback2008automated}, CUB \cite{WahCUB_200_2011} and ISIC-2016 \cite{gutman2016skin} datasets.
%since all compared methods except GrabCut \cite{rother2004grabcut} require a training set with a distribution identical to the test set to train their models while our pre-trained SC-VAE$^\curlywedge$ model can segment an image without the need for fine-tuning on Flower \cite{nilsback2008automated} and CUB \cite{WahCUB_200_2011} datasets.
Interestingly, our model performed better than most of the baselines, which used more sophisticated approaches for segmentation, such as adversarial training \cite{benny2020onegan, chen2019unsupervised, he2022ganseg, yu2021unsupervised, ding2022comgan} and attention mechanisms \cite{locatello2020object}. 
%We believe that the reason that SC-VAE$^\curlywedge$ outperformed the baselines is that SC-VAE$^\curlywedge$ can learn meaningful and distinctive sparse representations, which allows for learning relations between patches. 
We attribute the superior performance of SC-VAE$^\curlywedge$ over the baselines to its capability to learn meaningful and distinct sparse representations. This ability enables SC-VAE$^\curlywedge$ to learn relationships between patches effectively.
Several qualitative results of SC-VAE$^\curlywedge$ on Flower \cite{nilsback2008automated}, CUB \cite{WahCUB_200_2011} and ISIC-2016 \cite{gutman2016skin} datasets are shown in Figure \ref{figure:UIS_flowers}. More qualitative results can be found in the supplementary material.
%This is in contrast to most of the baselines, which use more sophisticated approaches for segmentation, such as adversarial training \cite{ chen2019unsupervised, he2022ganseg, savarese2021information, savarese2021information, yu2021unsupervised, ding2022comgan} and self-supervised learning \cite{hung2019scops}. The reason that SC-VAE outperformed the baselines is that SC-VAE can learn meaningful and distinctive sparse representations, which allows for learning relations between patches.\\
\begin{figure}[tbp]
\centering
\includegraphics[width=7.8cm]{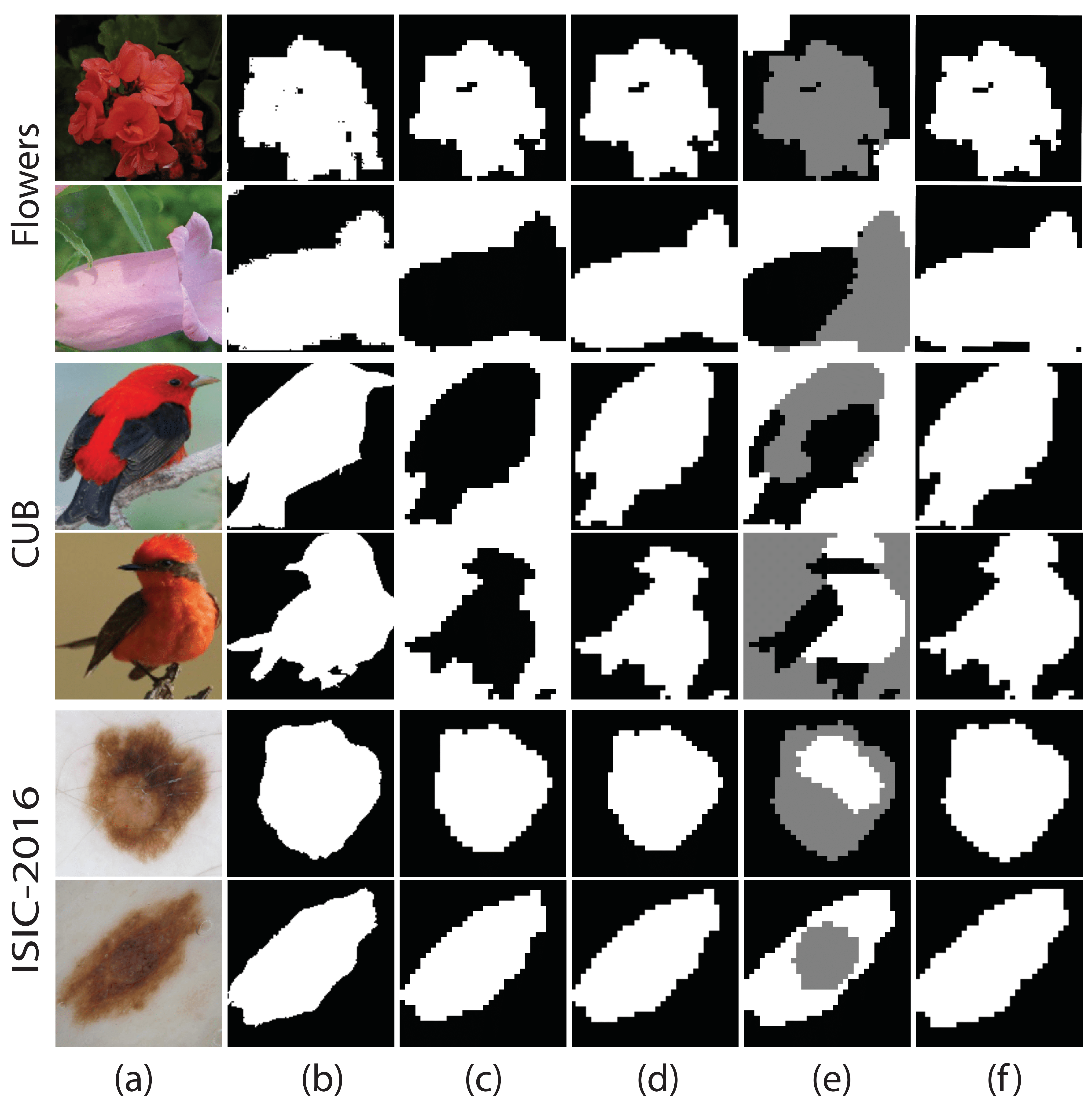}
\caption{Unsupervised image segmentation results of SC-VAE$^\curlywedge$ on Flowers \cite{nilsback2008automated}, CUB \cite{WahCUB_200_2011} and ISIC-2016 \cite{gutman2016skin}. \textit{Left to right:} (a) input image. (b) ground truth mask. (c) and (e) segmentation results by clustering sparse code vectors per image into $2$ or $3$ clusters using a spectral clustering algorithm. (d) and (f) boundary connectivity information \cite{zhu2014saliency} was used to decide the foreground and the background.}
\label{figure:UIS_flowers}
\vspace{-2pt}
\end{figure}

\begin{table}[!htp]
\centering
\caption{Unsupervised segmentation results on Flowers \cite{nilsback2008automated} and CUB \cite{WahCUB_200_2011}, measured in terms of IoU and DICE scores. SC-VAE $^\curlywedge$ was compared with state-of-the-art unsupervised and weakly supervised segmentation methods.
$\divideontimes$ denotes a GAN-based model.
OneGAN$\diamond$ is a weakly supervised baseline that relies on clean backgrounds as additional input. The best results across different metrics and datasets are highlighted in bold.} 
%\sisetup{detect-all}
\NewDocumentCommand{\B}{}{\fontseries{b}\selectfont}
\resizebox{0.85\linewidth}{!}{%
\begin{tabular}{
  @{}
  l
  S[table-format=1.2]
  S[table-format=1.2]|
  S[table-format=1.2]
  S[table-format=1.2]|
  S[table-format=1.2]
  S[table-format=1.2]
  @{}
}
\toprule
& \multicolumn{2}{c}{Flowers} & \multicolumn{2}{c}{CUB} & \multicolumn{2}{c}{ISIC-2016} \\
\cmidrule(lr){2-3} \cmidrule(lr){4-5} \cmidrule(lr){6-7}
Methods & { IoU $\uparrow$ } & { DICE $\uparrow$ } & { IoU $\uparrow$ } & { DICE $\uparrow$ } & { IoU $\uparrow$ } & { DICE $\uparrow$ }\\
\midrule
GrabCut \cite{rother2004grabcut} & {$69.2$} & {$79.1$} &  {$36.0$} & {$48.7$} & {-} & {-}  \\
W-Net \cite{xia2017w} & {$74.3$} & {$83.0$} &  {$24.8$} & {$38.9$} &  {$31.8$} & {$44.4$}  \\
ReDO$^{\divideontimes}$ \cite{chen2019unsupervised}  & {$77.8$} & {$85.3$}  & {$43.5$} & {$56.8$} & {$20.1$} & {$29.1$}  \\
IODINE \cite{greff2019multi}  & {$32.6$} & {$46.0$}  & {$30.9$} & {$44.6$}  & {$31.9$} & {$42.4$} \\
PerturbGAN$^\divideontimes$ \cite{bielski2019emergence}  & {-} & {-}  & {$38.0$} & {-} & {-} & {-}\\
OneGAN$\diamond^{\divideontimes}$ \cite{benny2020onegan}  & {-} & {-}  & {$55.5$} & {$69.2$} & {-} & {-}\\
Slot-Attn \cite{locatello2020object}  & {$32.9$} & {$45.7$}  & {$35.6$} & {$51.5$} & {$22.2$} & {$32.8$} \\
IEM \cite{savarese2021information}   & {$76.8$} & {$84.6$}  & {$52.2$} & {$66.0$} & {$\textbf{65.5}$} & {$\textbf{75.1}$}  \\
DRC \cite{yu2021unsupervised} &  {$46.9$} & {$60.8$}  & {$56.4$} & {$70.9$} & {$44.1$} & {$56.4$} \\
%GANSeg$^\divideontimes$ \cite{he2022ganseg} &  {$73.9$} & {-}  & {$\textbf{61.0}$} & {$\textbf{73.2}$}  \\
DS-ComGAN$^\divideontimes$ \cite{ding2022comgan} &  {$76.9$} & {$83.1$}  & {$\textbf{60.7}$} & {$71.3$} & {$43.9$} & {$57.8$} \\
\midrule
SC-VAE$^\curlywedge$ (2 classes) &  {$\textbf{81.2}$} & {$\textbf{88.5}$}   & {$53.2$} & {$67.6$} & {$61.6$} & {$70.7$} \\
SC-VAE$^\curlywedge$ (3 classes) & {$67.3$} & {$78.6$} & {$58.2$} & {$\textbf{72.1}$} & {$60.8$} & {$71.5$}\\
\bottomrule
\end{tabular}}
\label{tbl_un}
\end{table}

\begin{figure}[tbp]
\centering
\includegraphics[width=8cm]{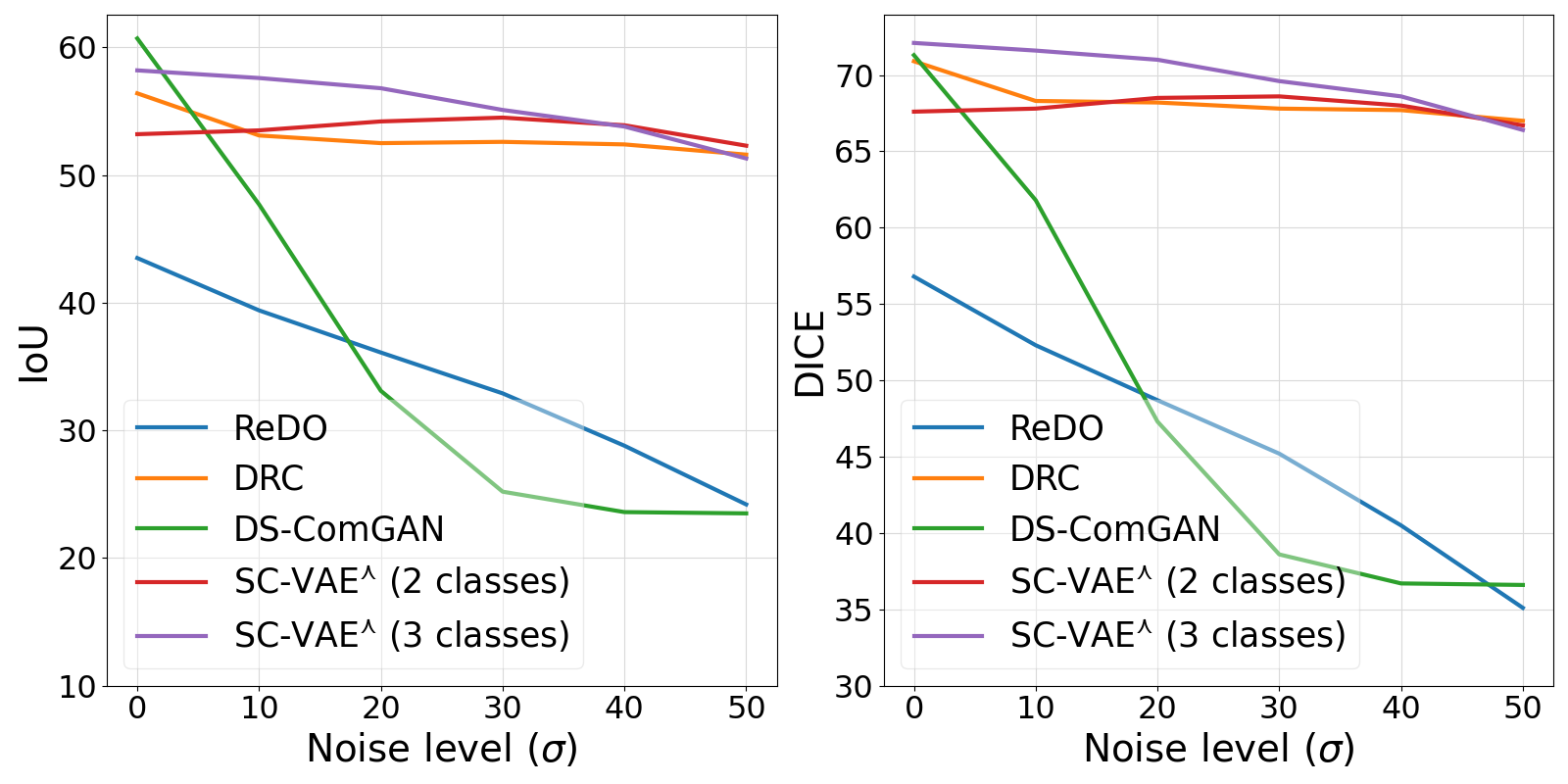}
\caption{Noise robustness analysis. We show the unsupervised segmentation results of SC-VAE$^\curlywedge$ and three baselines on CUB \cite{WahCUB_200_2011}  at different levels of noise $\sigma$.}
\label{figure:NRA}
\vspace{-2pt}
\end{figure}

\noindent
\textbf{Noise robustness analysis.} 
%Learnable ISTA was widely used to perform image denoising tasks. We exaimne if the pretrained SC-VAE$^\curlywedge$ has noise robustness property towards unsupervised image segmentation.We add i.i.d. Gaussian noise with zero mean and a specified level of noise $\sigma$ to images and compared with the performance of pretrained models including ReDo, DRC and DS-ComGAN. As is shown in Figure. \ref{figure:NRA}, SC-VAE$^\curlywedge$ and DRC is robust to Gaussian noise.
Sparse coding has been extensively applied for image denoising tasks. We investigate whether the pretrained SC-VAE$^\curlywedge$ exhibits resilience to noise in the unsupervised image segmentation task. To evaluate this, we added independent and identically distributed (i.i.d.) Gaussian noise with zero mean and various levels of noise $\sigma$ to each image in the test set of CUB dataset and compared its performance with three  baselines  (ReDO \cite{chen2019unsupervised}, DRC \cite{yu2021unsupervised}, and DS-ComGAN \cite{ding2022comgan}), which have been pretrained on the training set of CUB dataset. The performance of other baselines was not shown due to the unavailability of pretrained models. As depicted in Figure \ref{figure:NRA}, both SC-VAE$^\curlywedge$ and DRC \cite{yu2021unsupervised} demonstrated robustness against Gaussian noise whereas the performance of ReDO \cite{chen2019unsupervised} and DS-ComGAN \cite{ding2022comgan} noticeably declined as the noise level increased.
DRC \cite{yu2021unsupervised} includes Total Variation norm \cite{rudin1992nonlinear} within its loss function, which potentially explains the robustness to noise it exhibited.
%while the performance of ReDo\cite{chen2019unsupervised} and DS-ComGAN \cite{ding2022comgan} decrease significantly with the increasing of noise level.

\subsection{Ablation Study}
\noindent
We conducted an ablation study to analyze the influence of the number of rollout steps ($s$) in LISTA on the reconstruction ability of SC-VAE.  We trained the SC-VAE$^\curlyvee$ model ($d=4$) with different $s$ and reported the quantitative results in Table \ref{ablation_16x16}.
The sparsity of the learned sparse code vectors was calculated using the Hoyer metric \cite{hoyer2004non}. We reported the average sparsity and four reconstruction metrics among the validation set of FFHQ dataset. Sparsity first decreased and then increased as $s$ increased. Lower sparsity corresponds to better reconstruction outcomes.
\begin{table}[htbp]
\centering
\caption{Quantitative reconstruction results of SC-VAE ($d=4$) with different number of rollout steps $s$.} 
\resizebox{0.9\linewidth}{!}{
\begin{tabular}{c|c|c|c|c|c}
  \toprule
  Model  &  Sparsity & PSNR $\uparrow$ & SSIM $\uparrow$ & LPIPS $\downarrow$ & rFID $\downarrow$ \\
  \midrule
  SC-VAE ($s=1$)  &  $89.5\%$ & $27.30$ & $0.7603$ & $0.3039$ & $71.17$ \\
  SC-VAE ($s=5$)  &  $71.9\%$ & $31.13$ & $0.8759$ & $0.1279$ & $30.07$ \\
  SC-VAE ($s=10$) &  $75.1\%$ & $31.05$ & $0.8771$ & $0.1261$ & $29.50$ \\
  SC-VAE ($s=15$) &  $74.9\%$ & $31.41$ & $0.8815$ & $0.1155$ & $28.11$ \\
  SC-VAE ($s=20$) &  $80.9\%$ & $30.66$ & $0.8656$ & $0.1451$ & $32.87$ \\
  SC-VAE ($s=25$) &  $81.5\%$ & $30.59$ & $0.8693$ & $0.1373$ & $30.45$ \\
  SC-VAE ($s=30$) &  $83.2\%$ & $29.71$ & $0.8369$ & $0.1925$ & $41.89$ \\
  \bottomrule
\end{tabular}}
\label{ablation_16x16}
\vspace{-3pt}
\end{table}

\section{Conclusion}
%In this paper, we proposed a VAE variant, termed SC-VAE.
%The proposed method leverages an unrolled sparse coding learning algorithm to obtain latent sparse representations for the input image. The SC-VAE enabled the gradient flow through the dictionary and overcame the shortcomings of existing VAE models. In the experiments, we demonstrated that SC-VAE performed favorably compared to popular baseline methods in image reconstruction on two benchmark datasets. Additionally, 
%we demonstrated that similar image patches correspond to similar learned sparse codes. This allowed us to perform unsupervised image segmentation. The demonstrated advantages of SC-VAE compared to standard VAE approaches may enable better performance in downstream tasks, such as object recognition, image segmentation, and image retrieval. In future research, we will explore the potential of combining SC-VAE with a transformer model to investigate its ability to generate high-quality images. Additionally, we intend to expand our research into (weakly) supervised scenarios.
In this paper, we introduced a novel variant of VAE, which we refer to as SC-VAE. Our approach harnesses a learned ISTA algorithm to learn latent sparse representations for input images. The utilization of SC-VAE facilitates the smooth flow of gradients through the network, effectively addressing limitations present in existing VAE models. Through our experiments, we showcased the superior performance of SC-VAE compared to well-established baseline methods in image reconstruction. 
Furthermore, we illustrated that SC-VAE can generate new images through manipulating and interpolating sparse code vectors. Moreover, 
SC-VAE's ability to learn relationship between image patches enabled us to perform unsupervised image segmentation, coupled with its resilience to noise.
%SC-VAE's ability to learn relationship between image patches using sparse codes allows us to perform unsupervised image segmentation associated with its noise robustness property. %We also demonstrate the noise robustness property of SC-VAE for unsupervised image segmentation. 
%The proposed SC-VAE has the potential to enhance performance in downstream tasks such as image classification, image retrieval and object detection.
%For future work, we plan to explore the potential of combining SC-VAE with a Transformer model to assess its capacity for generating high-fidelity images. %Additionally, we aim to broaden the scope of our research by delving into (weakly) supervised scenarios.

\newpage
% \clearpage
% \pagebreak
% \onecolumn % 

\clearpage
\maketitlesupplementary
The supplementary material for our work  \textit{SC-VAE: Sparse Coding-based Variational Autoencoder with Learned ISTA} is structured as follows:
%Sec. \ref{section1_s} provides the detailed information of the encoder and decode architecture of the SC-VAE model. 
%Sec. \ref{section2_s} shows the visualization of the dictionary atoms.
%Sec. \ref{section3_s} shows the training loss on the ImageNet dataset with different number of downsampling (upsampling) blocks ($d$) in the encoder (decoder) of the SC-VAE model.
%Sec. \ref{section4_s} shows the visualization results of an unofficial implementation of VIT-VQGAN \cite{yu2021vector}. 
%Sec. \ref{section5_s} shows additional manipulation and interpolation results on FFHQ dataset. 
%Sec. \ref{section6_s} shows additional image patches clustering results on FFHQ and ImageNet datasets. 
%Sec. \ref{section7_s} shows additional unsupervised image segmentation results.
Section \ref{section1_s} details the encoder and decoder architecture of the SC-VAE model. In Section \ref{section2_s}, the dictionary atoms are visualized. In Section \ref{section3_s}, we provide the training losses on the ImageNet dataset when varying the number of downsampling (upsampling) blocks ($d$) in the encoder (decoder) of the SC-VAE model. In Section \ref{section4_s}, the visualized reconstruction results of an unofficial implementation of VIT-VQGAN \cite{yu2021vector} are provided. 
We provide  additional manipulation and interpolation results on the FFHQ dataset in Section \ref{section5_s}, while  additional clustering results of image patches on both FFHQ and ImageNet  are provided in Section \ref{section6_s}. Supplementary unsupervised image segmentation results are given in Section \ref{section7_s}.

%Additional results on image patches clustering and unsupervised image segmentation on FFHQ and ImageNet datasets are then presented in Sec. 2 and Sec. 3, respectively.
\setcounter{section}{0}

\section{The Encoder and Decoder Architecture of SC-VAE} \label{section1_s}
The SC-VAE model's encoder and decoder architecture mirrors that of VQGAN \cite{esser2021taming}. Details about the architecture are provided in Table \ref{figure:encoder_decoder}.
%The encoder and decoder architecture in the SC-VAE model are the same as the architecture used in VQGAN \cite{esser2021taming}, which is described in Table \ref{figure:encoder_decoder}. 
$H$, $W$ and $C$ denote the height, width
and the number of channels of an input image, respectively.
$C'$ and $C''$ represent the number of channels of the feature maps that are produced as outputs by the intermediate layers of the encoder and decode network.
In our experiment, $C'$ and $C''$ were set to $128$ and $512$, respectively. $n$ denotes the number of dimensions of each latent representation, which was set to $256$.
The variable $d$ represents the number of blocks used for downsampling and upsampling. Therefore, we can calculate the height ($h$) and width ($w$) of the encoder's output feature maps by dividing the height ($H$) and width ($W$) of input images by $2$ raised to the power of $d$.

\begin{table}[thbp!]
\centering
\caption{High-level architecture of the encoder and decoder of the SC-VAE model. $H$, $W$, and $C$ refer to the height, width, and the number of channels of an input image. 
$C'$ and $C''$ represent the number of channels of the feature maps from intermediate layers in the encoder and decoder networks. $n$ denotes the number of dimensions of each latent representation, while $d$ represents the number of downsampling (upsampling) blocks. Note that $h=\frac{H}{2^{d}}$, $w=\frac{W}{2^d}$.} 
\resizebox{1\linewidth}{!}{%
\begin{tabular}{c|c}
  \toprule
   &  $x\in \mathbb{R}^{H\times W\times C} $\\
   &  2D Convolution $\rightarrow \mathbb{R}^{H\times W\times C'}$\\
   &  $d \times$\{Residual Block, Downsample Block\} $\rightarrow \mathbb{R}^{h\times w\times C''}$\\
   &  Residual Block $\rightarrow \mathbb{R}^{h\times w\times C''}$\\
  Encoder &  Non-Local Block $\rightarrow \mathbb{R}^{h\times w\times C''}$\\
   &  Residual Block $\rightarrow \mathbb{R}^{h\times w\times C''}$\\
   &  Group Normalization \cite{wu2018group} $\rightarrow \mathbb{R}^{h\times w\times C''}$ \\
   &  Swish Activation Function \cite{ramachandran2017searching} $\rightarrow \mathbb{R}^{h\times w\times C''}$\\
   &  2D Convolution $\rightarrow E(x) \in \mathbb{R}^{h\times w\times n}$\\
  \midrule
   & $\tilde{E}(x)\in \mathbb{R}^{h\times w\times n} $  \\
   &  2D Convolution $\rightarrow \mathbb{R}^{h\times w\times C''}$  \\
   &  Residual Block $\rightarrow \mathbb{R}^{h\times w\times C''}$\\
    & Non-Local Block $\rightarrow \mathbb{R}^{h\times w\times C''}$\\
  Decoder & Residual Block $\rightarrow \mathbb{R}^{h\times w\times C''}$\\
    & $d\times$\{Residual Block, Upsample Block\} $\rightarrow \mathbb{R}^{H\times W\times C'}$\\
   & Group Normalization \cite{wu2018group} $\rightarrow \mathbb{R}^{H\times W\times C'}$\\
    & Swish  Activation Function \cite{ramachandran2017searching}
    $\rightarrow \mathbb{R}^{H\times W\times C'}$\\
    & 2D Convolution $\rightarrow G(\tilde{E}(x)) \in \mathbb{R}^{H\times W\times C}$\\
  \bottomrule
\end{tabular}}
\label{figure:encoder_decoder}
\end{table}

\section{Visualization of Dictionary Atoms}
\label{section2_s}
\begin{figure}[tbp]
\centering
\includegraphics[width=8cm]{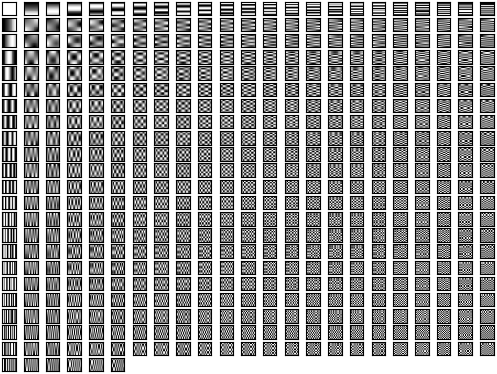}
\caption{$512$ atoms of the Discrete Cosine Transform (DCT) dictionary. All atoms were reshaped into a $16 \times 16$ image.}
\label{figure:dictionary_visualization}
\end{figure}

Figure \ref{figure:dictionary_visualization} demonstrates the $512$ columns (atoms) of the pre-determined Discrete Cosine Transform (DCT) dictionary. Each atom is of dimension $256$, which corresponds to the size of $16 \times 16$ images when shaped.
%We reshape all atoms into an image with a $16\times 16$ resolution.

\section{Training Losses}  \label{section3_s}
\begin{figure}[tbp]
\centering
\includegraphics[width=7.5cm]{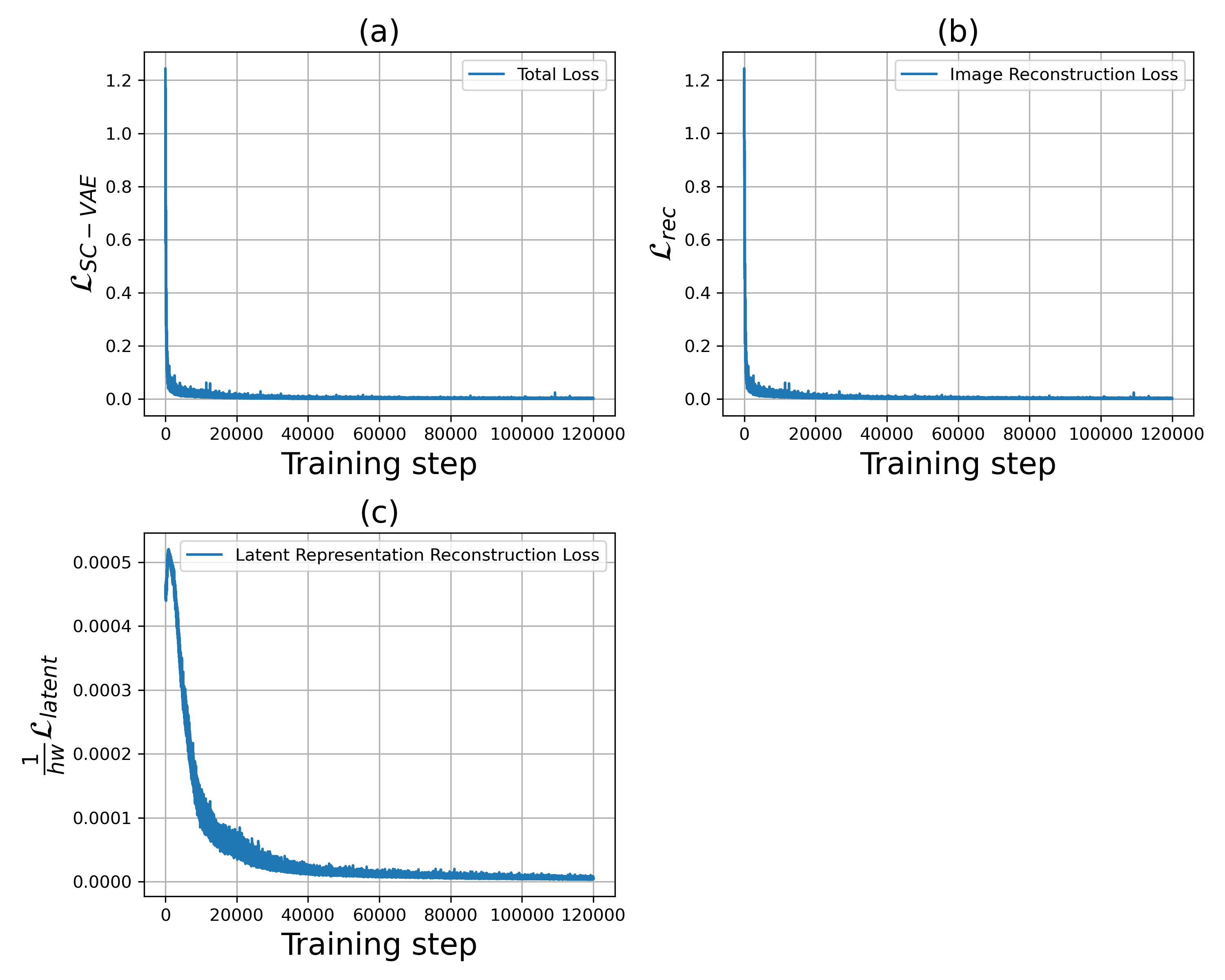}
\caption{The training losses over $120,000$ training steps on the ImageNet dataset. The number of  downsampling (upsampling) blocks ($d$) in the encoder (decoder) of the SC-VAE model was set to $3$ and the height ($h$) and width ($w$) of latent representations were $32$. (a) Total loss $\mathcal{L}_{SC-VAE}$. (b) Image reconstruction loss $\mathcal{L}_{rec}$. (c)The mean of latent representations reconstruction loss $\frac{1}{hw}\mathcal{L}_{latent}$.}
\label{figure:TLImagenet32x32}
\end{figure}

\begin{figure}[tbp]
\centering
\includegraphics[width=7.5cm]{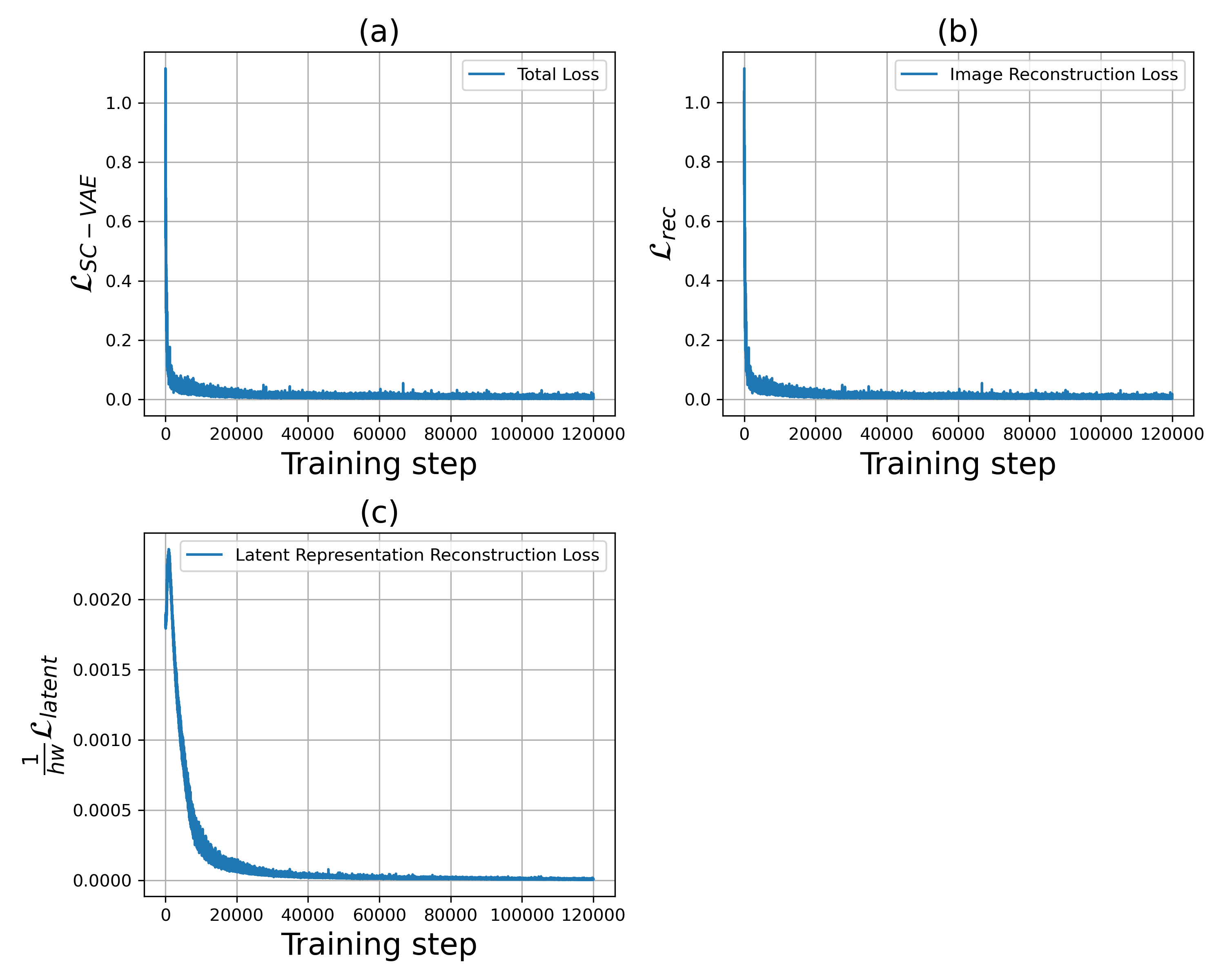}
\caption{The training losses over $120,000$ training steps on the ImageNet dataset. The number of  downsampling (upsampling) blocks ($d$) in the encoder (decoder) of the SC-VAE model was set to $4$ and the height ($h$) and width ($w$) of latent representations were $16$. (a) Total loss $\mathcal{L}_{SC-VAE}$. (b) Image reconstruction loss $\mathcal{L}_{rec}$. (c) The mean of latent representations reconstruction loss $\frac{1}{hw}\mathcal{L}_{latent}$.}
\label{figure:TLImagenet16x16}
\end{figure}

\begin{figure}[tbp]
\centering
\includegraphics[width=7.5cm]{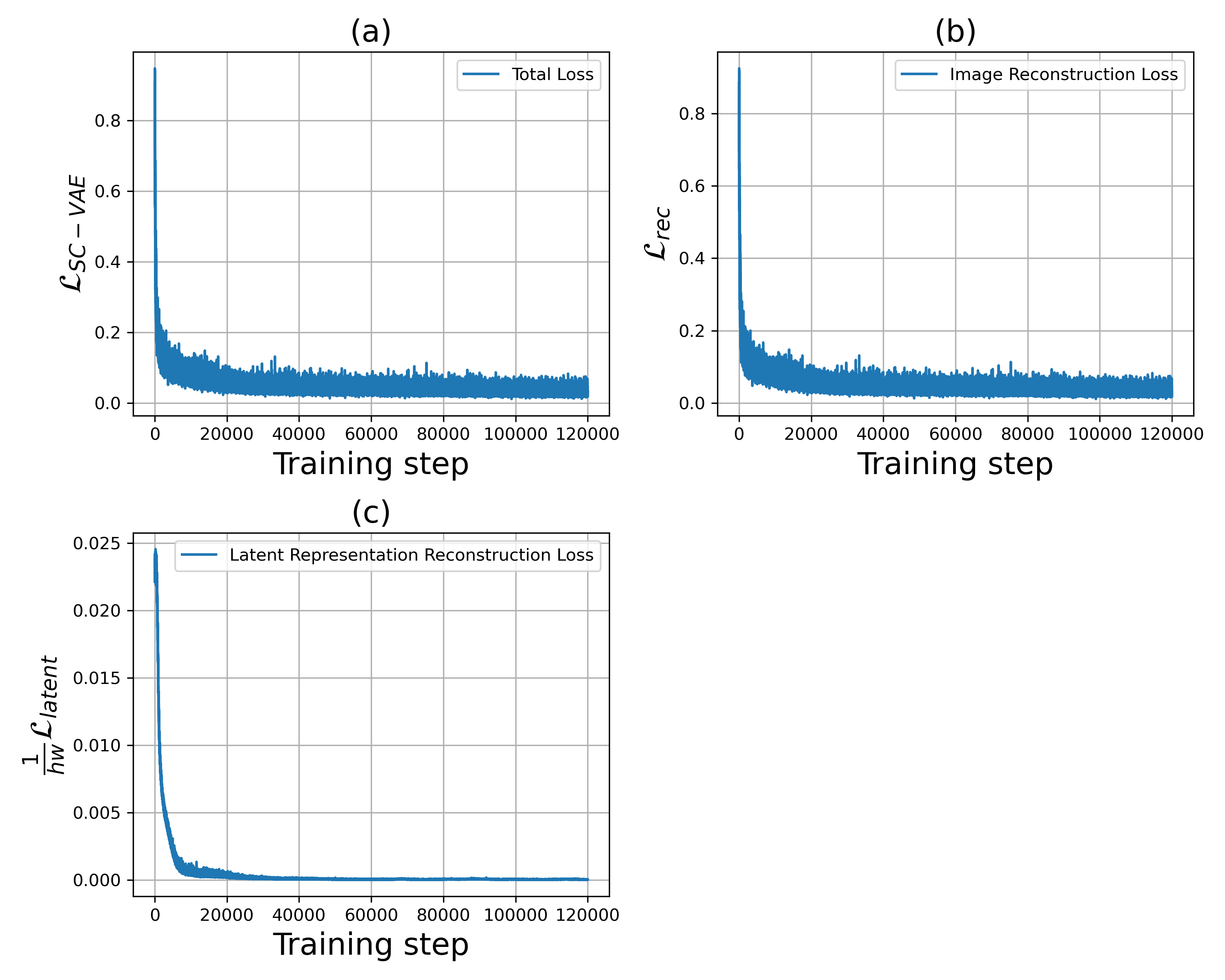}
\caption{The training losses over $120,000$ training steps on the ImageNet dataset. The number of  downsampling (upsampling) blocks ($d$) in the encoder (decoder) of the SC-VAE model was set to $6$ and the height ($h$) and width ($w$) of latent representations were $4$. (a) Total loss $\mathcal{L}_{SC-VAE}$. (b) Image reconstruction loss $\mathcal{L}_{rec}$. (c) The mean of latent representations reconstruction loss $\frac{1}{hw}\mathcal{L}_{latent}$.}
\label{figure:TLImagenet4x4}
\end{figure}

\begin{figure}[tbp]
\centering
\includegraphics[width=7.5cm]{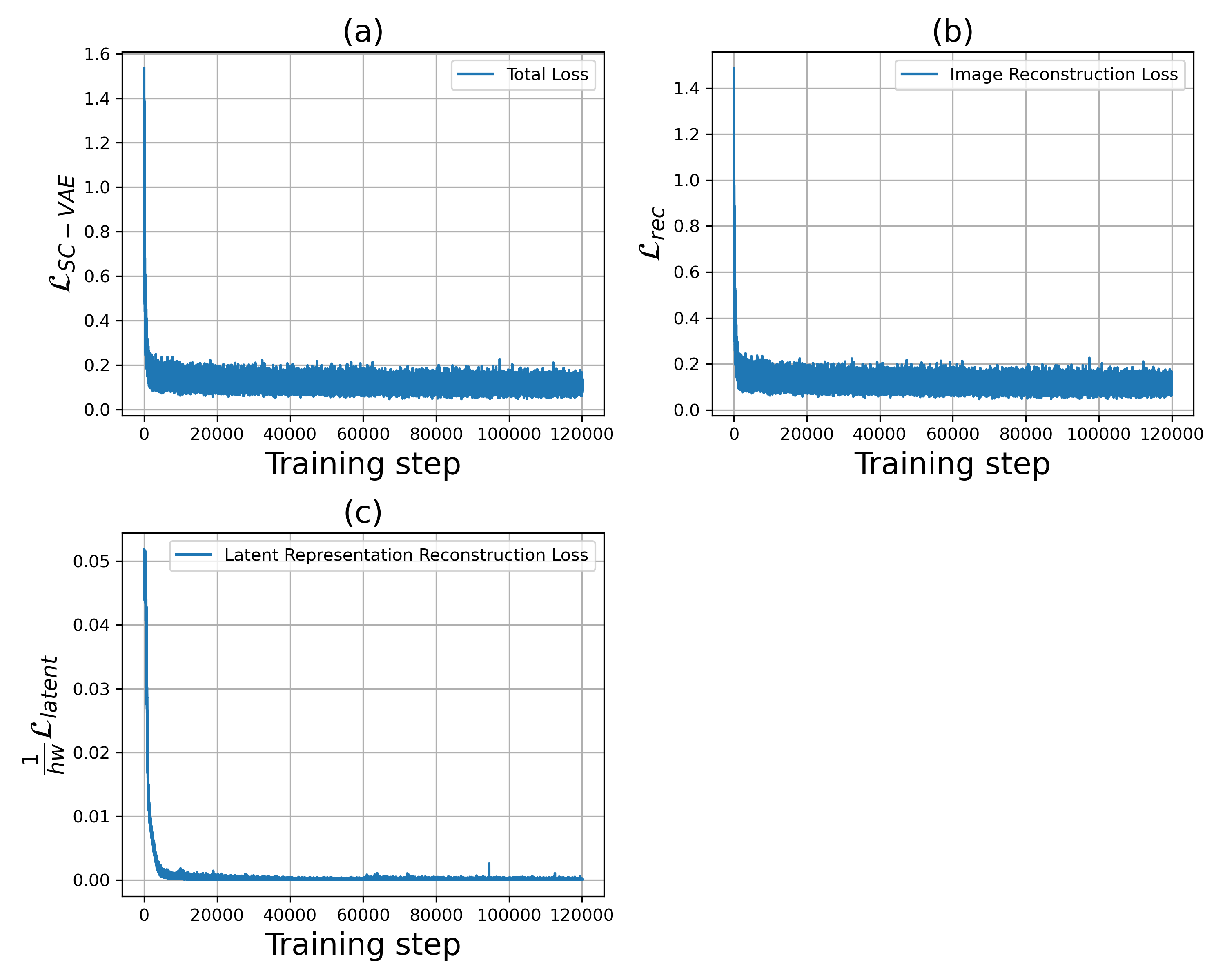}
\caption{The training losses over $120,000$ training steps on the ImageNet dataset. The number of  downsampling (upsampling) blocks ($d$) in the encoder (decoder) of the SC-VAE model was set to $8$ and the height ($h$) and width ($w$) of latent representations were $1$. (a) Total loss $\mathcal{L}_{SC-VAE}$. (b) Image reconstruction loss $\mathcal{L}_{rec}$. (c) The mean of latent representations reconstruction loss $\frac{1}{hw}\mathcal{L}_{latent}$.}
\label{figure:TLImagenet1x1}
\end{figure}

%Training losses of inherent noises around the 140th epoch under different auxiliary dataset sizes (K)
Figures \ref{figure:TLImagenet32x32}, \ref{figure:TLImagenet16x16}, \ref{figure:TLImagenet4x4} and \ref{figure:TLImagenet1x1} show  the training losses over $120,000$ training steps on the
ImageNet dataset.
The number of downsampling (upsampling) blocks ($d$) in the encoder (decoder) of the SC-VAE model are $3, 4, 6$ and $8$, respectively.
%with the number of downsampling (upsampling) blocks ($d=3,4,6$ and $8$, respectively) in the encoder (decoder) of the SC-VAE model. 
%As is shown in these figures, the LISTA networks of the SC-VAE models converge to a fixed point no matter which downsampling (upsampling) block $d$ is used. However, SC-VAE suffer from image reconstruction when increasing $d$.
As depicted in these figures, the LISTA networks within the SC-VAE models consistently converge to a stable point regardless of the chosen downsampling (upsampling) block $d$. However, increasing $d$ leads to worse image reconstructions ($\mathcal{L}_{rec}$) in SC-VAE.

%\noindent
%\noindent\textbf{Learnbale ISTA.} The architecture of our Learnable ISTA network is shown in Table 2.
%\noindent\textbf{Attention Network for $\alpha$ Estimation.}  Our neural network architecture follows the backbone of PixelCNN++ [52], which is a U-Net [48] based on a Wide ResNet [72]. We replaced weight normalization [49] with group normalization [66] to make the implementation simpler. Our 32 × 32 models use four feature map resolutions (32 × 32 to 4 × 4), and our 256 × 256 models use six. All models have two convolutional residual blocks per resolution level and self-attention blocks at the 16 × 16 resolution between the convolutional blocks [6].

% \begin{table*}[!htbp]
% \centering
% \caption{High-level architecture of the Learnable ISTA of our SC-VAE. Note that $k$ is the number of the unfolded ISTA block.} 
% \begin{tabular}{c}
%   \toprule
%   Learnable ISTA \\
%   \midrule
%   $E(x)\in \mathbb{R}^{h\times w \times n} $ \\
%   Filter Matrix $\rightarrow \mathbb{R}^{h\times w\times K}$ \\
%   $k\times$\{Shrinkage Function, Mutual Inhibition Matrix, Addition Operator\} $\rightarrow \mathbb{R}^{h\times w\times K}$\\
%   Shrinkage function$\rightarrow Z\in \mathbb{R}^{h\times w\times K}$\\
%   \bottomrule
% \end{tabular}
% \end{table*}

\section{Image Reconstruction}  \label{section4_s}
\begin{figure}[tbp]
\centering
\includegraphics[width=7.0cm]{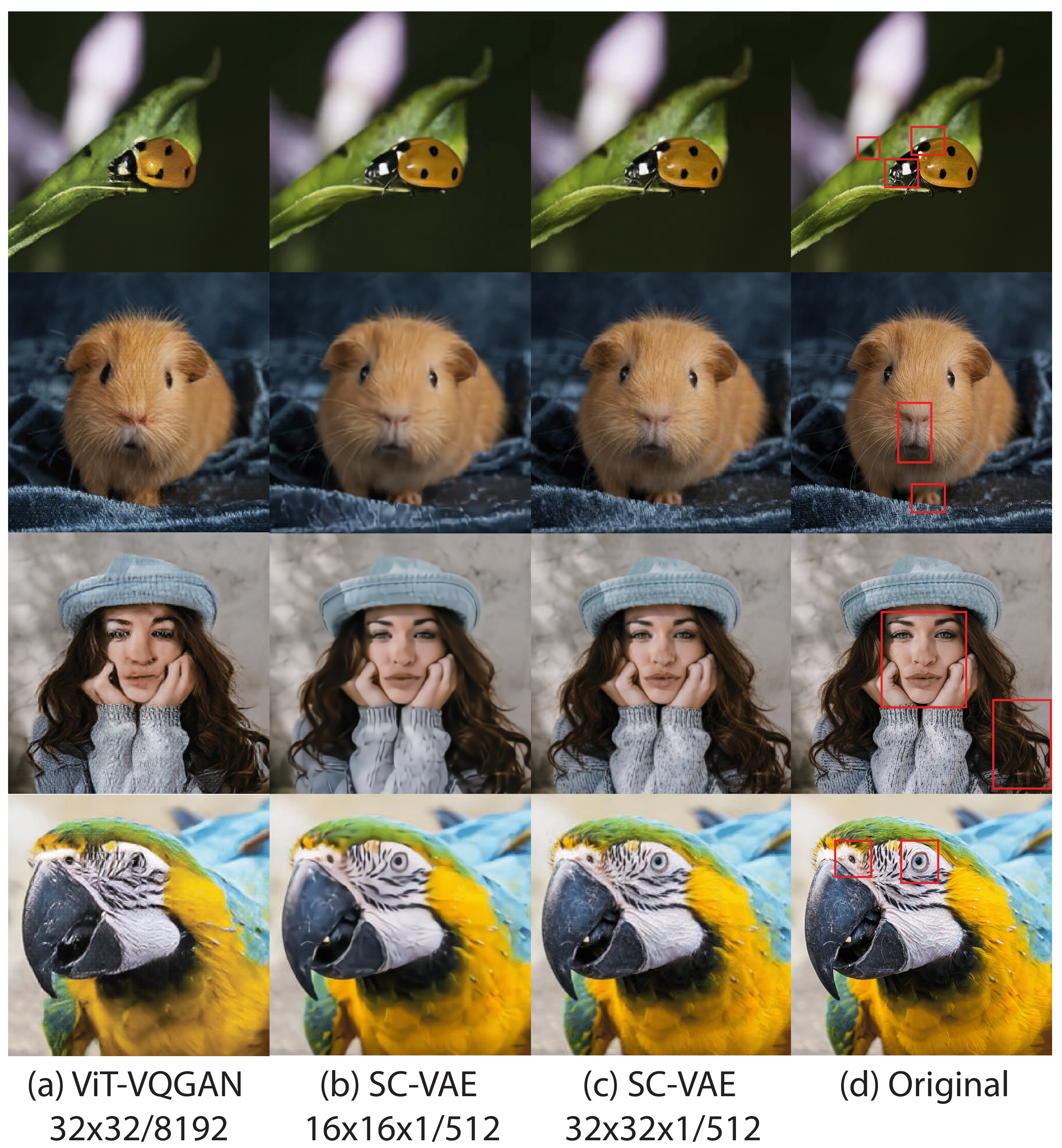}
\caption{Image reconstructions from an unofficial implementation of VIT-VQGAN \cite{yu2021vector} and the SC-VAE models trained
on ImageNet dataset. Original images in the top two rows are
from the validation set of ImageNet dataset. Two external images are shown in the last two rows to demonstrate the generalizability of different methods. The numbers denote the shape of
latent codes and the learned codebook (dictionary) size, respectively.
SC-VAE achieved improved image reconstruction compared to VIT-VQGAN \cite{yu2021vector}. Zoom in to see the details in the red square area.}
\label{figure:ViT-VQGAN_Visualization}
\end{figure}

Reconstruction results from unofficial implementation\footnote{https://github.com/thuanz123/enhancing-transformers} of VIT-VQGAN \cite{yu2021vector} are presented in Figure \ref{figure:ViT-VQGAN_Visualization}.
%Figures \ref{figure:ViT-VQGAN_Visualization} shows visualizations from unofficial implementation\footnote{https://github.com/thuanz123/enhancing-transformers} of VIT-VQGAN \cite{yu2021vector}. 
VIT-VQGAN \cite{yu2021vector} achieved visually appealing results. However, similar to VQ-GAN \cite{esser2021taming} and RQ-VAE \cite{lee2022autoregressive}, it faced challenges in accurately reconstructing intricate details and complex patterns.
%as VQ-GAN\cite{esser2021taming} and RQ-VAE\cite{lee2022autoregressive}. 
Additionally, its generalization performance was inferior to that of our model.

\section{Image Generation}  \label{section5_s}
\begin{figure}[tbp]
\centering
\includegraphics[width=7.0cm]{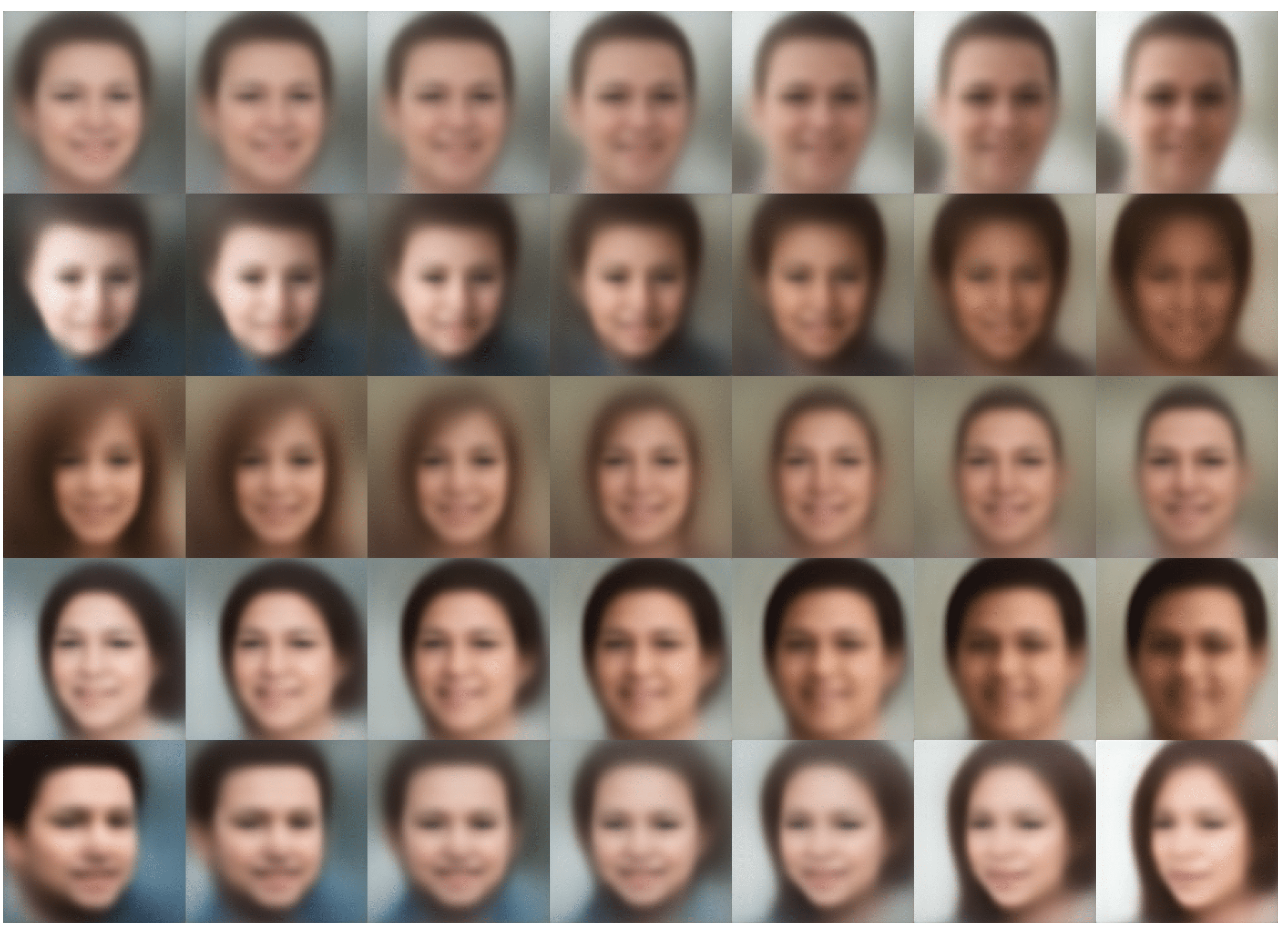}
\caption{Interpolation between the sparse code vectors of two samples from the SC-VAE$^{\dag}$ model trained on FFHQ.}
\label{figure:image_interpolation_supple}
\end{figure}

\begin{figure*}[tbp]
\centering
\includegraphics[width=14.5cm]{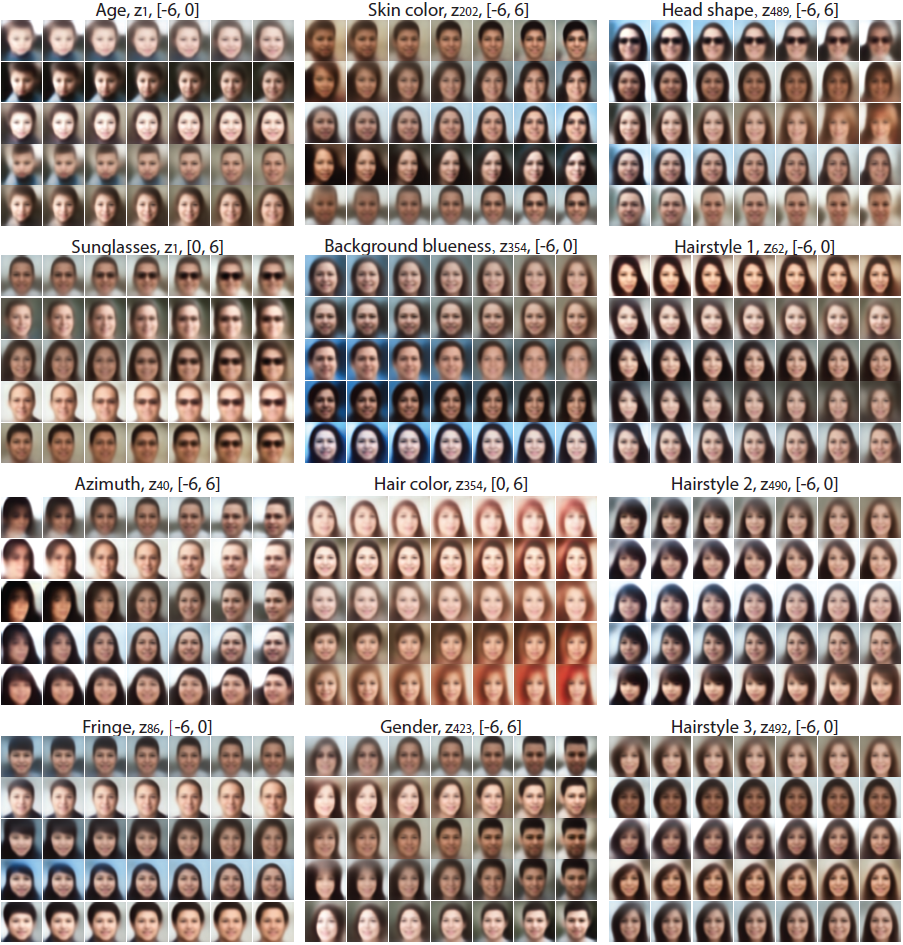}
\caption{Manipulating sparse code vectors on FFHQ. 
Each block contains five seed images used to infer the latent sparse code vector in the SC-VAE$^{\dag}$ model.
The disentangled attributes associated with the $i$-th component of a sparse code vector $z$ and a traversal range are shown on the top of each block.}
\label{figure:image_manipulation_supple}
\end{figure*}

Additional interpolation and manipulation results can be found in Figures \ref{figure:image_interpolation_supple} and \ref{figure:image_manipulation_supple}, respectively.

% \begin{figure}[tbp]
% \centering
% \includegraphics[width=8cm]{./Figures/IG_Age.png}
% \caption{IG-Age.}
% \label{figure:IG_Age}
% \end{figure}

% \begin{figure}[tbp]
% \centering
% \includegraphics[width=8cm]{./Figures/IG_sunglasses.png}
% \caption{IG-sunglasses.}
% \label{figure:IG_sunglasses}
% \end{figure}

% \begin{figure}[tbp]
% \centering
% \includegraphics[width=8cm]{./Figures/IG_Azimuth.png}
% \caption{IG-azimuth.}
% \label{figure:IG_azimuth}
% \end{figure}

% \begin{figure}[tbp]
% \centering
% \includegraphics[width=8cm]{./Figures/IG_Fringe.png}
% \caption{IG-Fringe.}
% \label{figure:IG_Fringe}
% \end{figure}

% \begin{figure}[tbp]
% \centering
% \includegraphics[width=8cm]{./Figures/IG_skin color.png}
% \caption{IG-skin color.}
% \label{figure:IG_skin color}
% \end{figure}

% \begin{figure}[tbp]
% \centering
% \includegraphics[width=8cm]{./Figures/image_interpolation_supple.png}
% \caption{Interpolation in the latent space between two samples from a model trained on FFHQ.}
% \label{figure:interpolation}
% \end{figure}

\section{Image Patches Clustering}  \label{section6_s}
\begin{figure*}[h!]
\centering
\includegraphics[width=16cm]{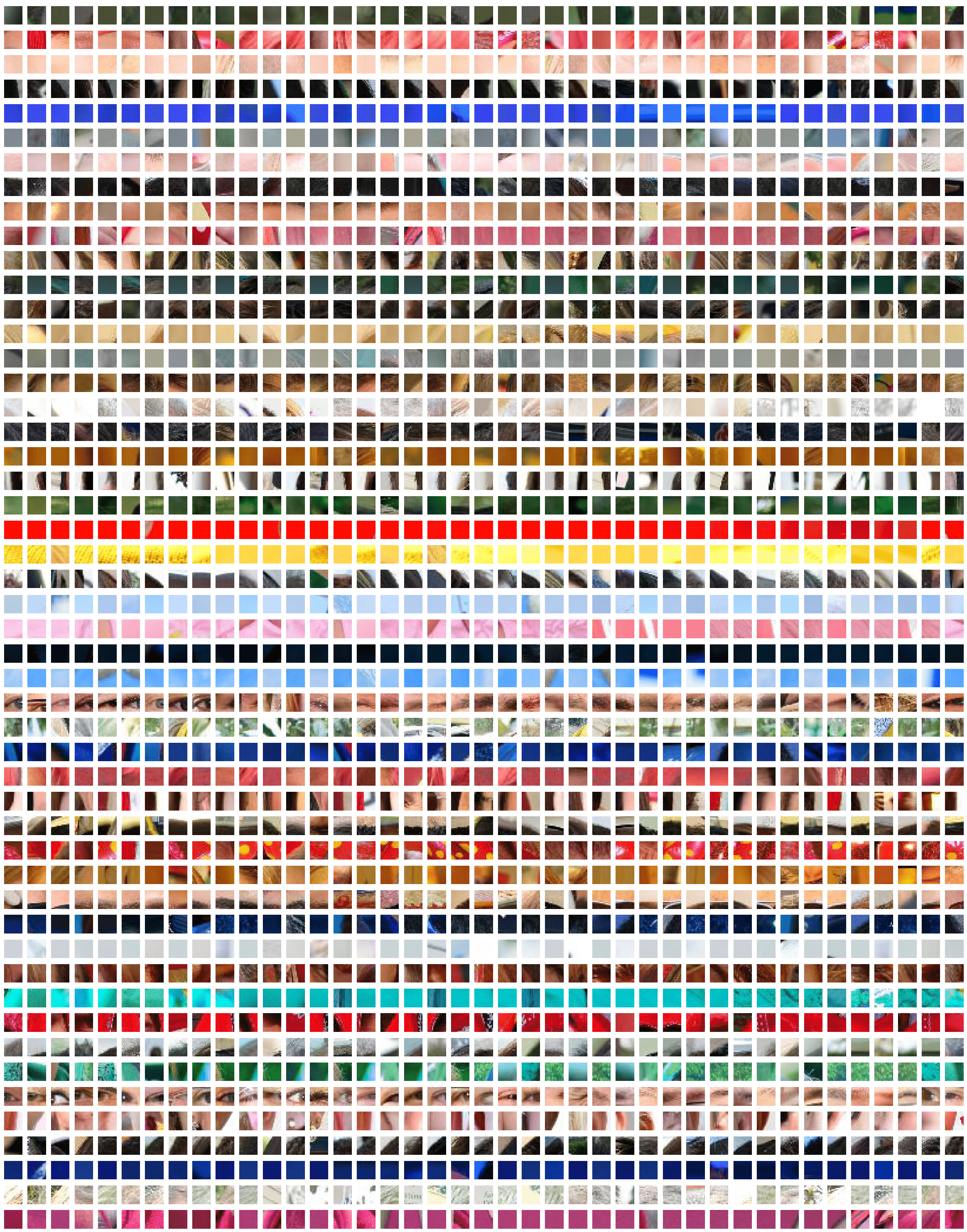}
\caption{50 randomly selected image patch clusters from the validation set of the FFHQ dataset generated by clustering the learned sparse code vectors of the pre-trained SC-VAE$^\curlyvee$ model
using the K-means algorithm. Each row represents one cluster. Image patches with similar patterns were grouped together.}
\label{figure:s1}
\end{figure*}

\begin{figure*}[h!]
\centering
\includegraphics[width=16cm]{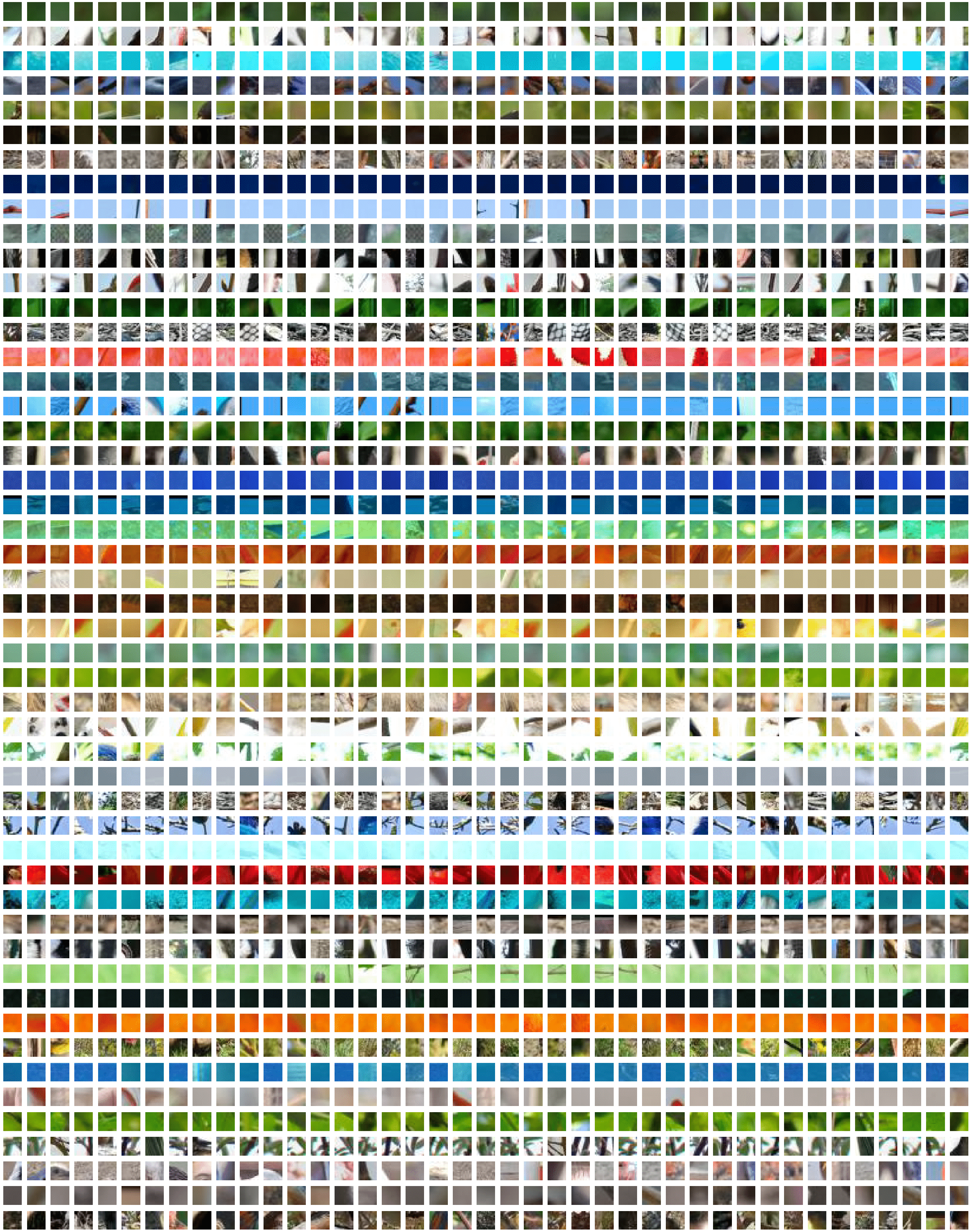}
\caption{50 randomly selected image patch clusters from the validation set of the ImageNet dataset generated by clustering the learned sparse code vectors of the pre-trained SC-VAE$^\curlyvee$ model
using the K-means algorithm. Each row represents one cluster. Image patches with similar patterns were grouped together.}
\label{figure:s2}
\end{figure*}

%Figures \ref{figure:s1} and \ref{figure:s2} exhibit more image patches clustering outcomes for the FFHQ and ImageNet datasets, respectively. 
Figures \ref{figure:s1} and \ref{figure:s2} showcase additional qualitative results of image patches clustering on FFHQ and ImageNet datasets, respectively.
These results were obtained utilizing the pre-trained SC-VAE$^\curlyvee$ model specific to each dataset with a downsampling block $d=4$.

% \begin{figure*}[h!]
% \centering
% \includegraphics[width=16cm]{./Figures/segmentation_ffhq_supple3.png}
% \caption{FFHQ.}
% \label{figure:5}
% \end{figure*}

\section{Unsupervised Image Segmentation} \label{section7_s}
\subsection{Qualitative Analysis on FFHQ and ImageNet}
\begin{figure*}[h!]
\centering
\includegraphics[width=16cm]{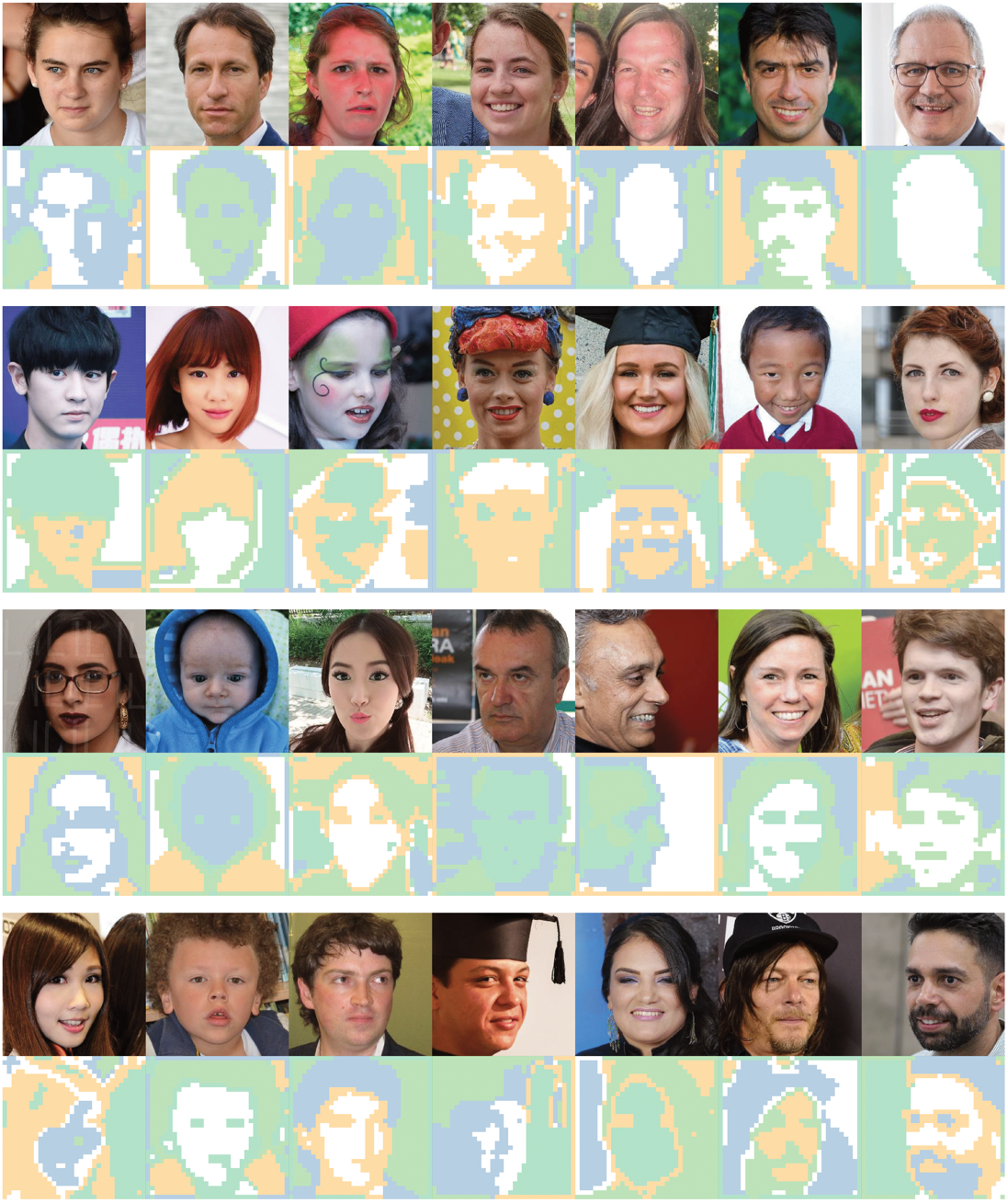}
%\caption{Additional unsupervised image segmentation results. Images are from the validation set of the FFHQ dataset.}
\caption{Additional unsupervised image segmentation results. These results were generated by grouping sparse code vectors into $5$ categories per image, utilizing the pre-trained SC-VAE$^{\curlywedge}$ model and the K-means algorithm. Images are from the validation set of the FFHQ dataset.}
\label{figure:s3}
\end{figure*}

\begin{figure*}[h!]
\centering
\includegraphics[width=16cm]{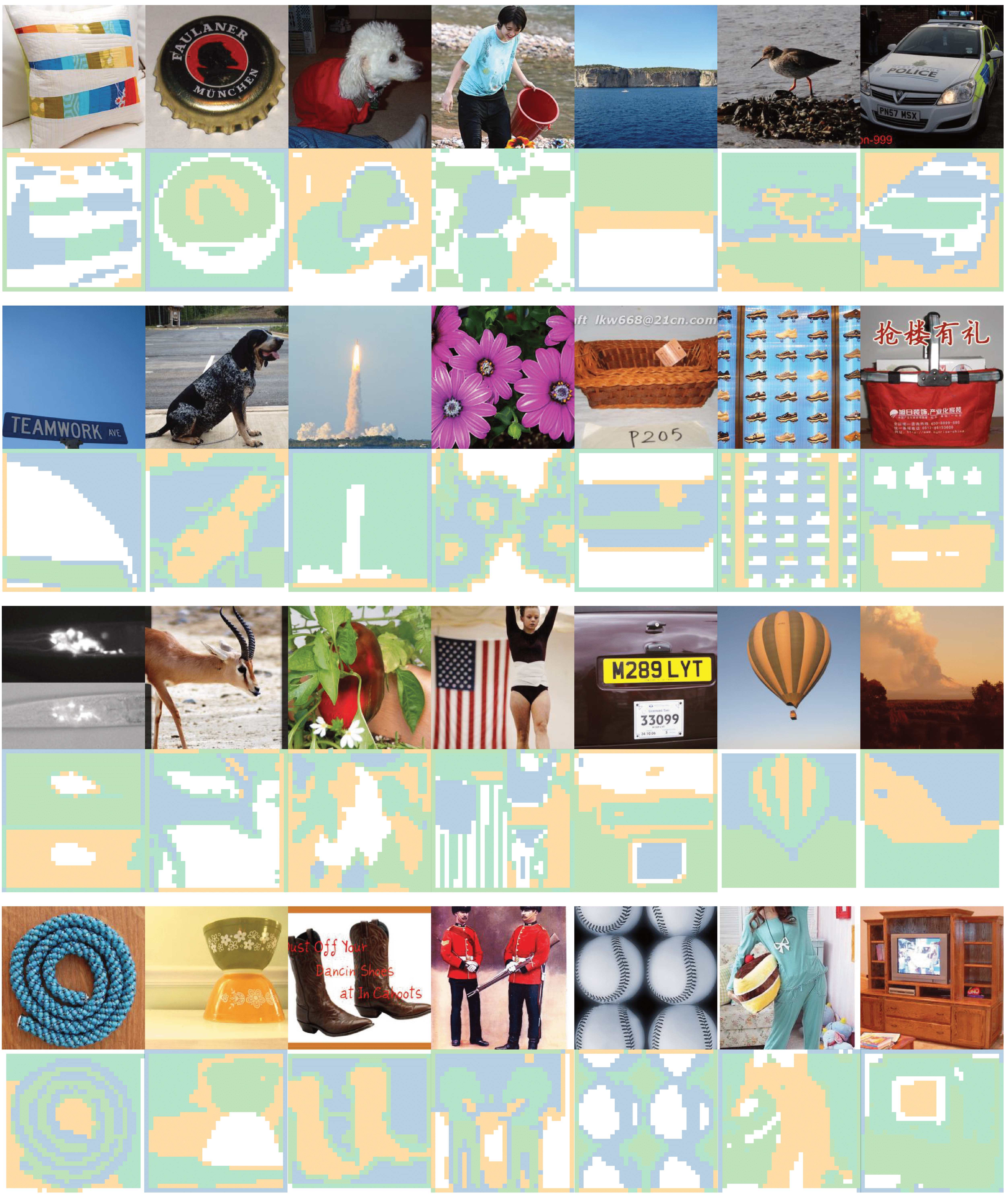}
%\caption{Additional unsupervised image segmentation results by applying K-means algorithm to cluster sparse code vectors per image into $5$ categories using the SC-VAE$^{\curlywedge}$ model. Images are from the validation set of the ImageNet dataset.}
\caption{Additional unsupervised image segmentation results. These results were generated by grouping sparse code vectors into $5$ categories per image, utilizing the pre-trained SC-VAE$^{\curlywedge}$ model and the K-means algorithm. Images are from the validation set of the ImageNet dataset.}
\label{figure:s4}
\end{figure*}

%Figures \ref{figure:s3} and \ref{figure:s4} contain additional qualitative unsupervised image segmentation results on FFHQ and ImageNet datasets, respectively. 
%We utilized two SCVAE models that were pre-trained on the training set of the FFHQ and ImageNet dataset, respectively. These models had a downsampling block of $d = 3$ and a sparsity penalty of $\lambda = 2$. 
%We employed two SC-VAE$^\curlywedge$ models that had been pre-trained on the training sets of the FFHQ and ImageNet datasets, respectively. These models had a downsampling block $d=3$.
Additional qualitative unsupervised image segmentation results on the FFHQ and ImageNet datasets can be found in Figures \ref{figure:s3} and \ref{figure:s4}, respectively. We utilized two SC-VAE$^\curlywedge$ models pre-trained on the training sets of FFHQ and ImageNet, each employing a downsampling block $d=3$.
\subsection{Quantitative comparisons to prior work}
\begin{figure*}[tbp]
\centering
\includegraphics[width=18cm]{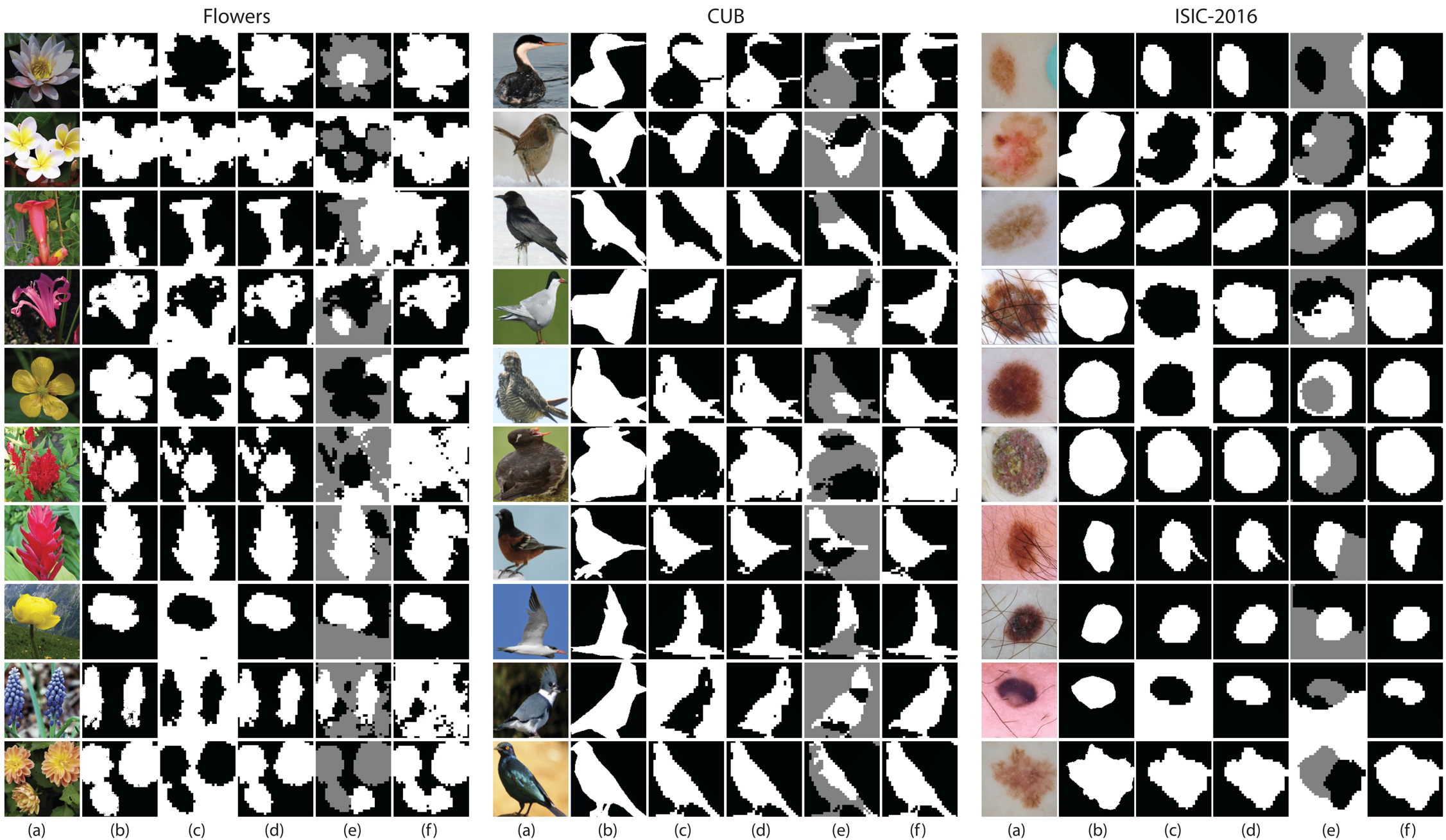}
\caption{Additional unsupervised image segmentation results on Flowers \cite{nilsback2008automated} (\textit{Left Panel}), CUB \cite{WahCUB_200_2011} (\textit{Middle Panel}) and ISIC-2016 \cite{gutman2016skin} (\textit{Right Panel}). (a) input image. (b) ground truth mask. (c) and (e) segmentation results by clustering sparse code vectors per image into $2$ or $3$ classes using a spectral clustering algorithm. (d) and (f) boundary connectivity information \cite{zhu2014saliency}
was used to decide the foreground and background.}
\label{figure:Flower_CUB}
\end{figure*}

%Figure \ref{figure:Flower_CUB} shows more qualitative results on  Flowers \cite{nilsback2008automated} and Caltech-UCSD Birds-200-2011 (CUB) \cite{WahCUB_200_2011}. Flowers \cite{nilsback2008automated} consists of $8,189$ images of $102$ classes of flowers, with segmentation masks obtained by an automated algorithm developed specifically for segmenting flowers in color photographs \cite{nilsback2007delving}. CUB \cite{WahCUB_200_2011} consists of $11,788$ images of $200$ classes of birds and segmentation masks. Flowers and CUB contain $1,020$ and $1,000$ test images, respectively.
%Figure \ref{figure:Flower_CUB} shows more qualitative results on  Flowers \cite{nilsback2008automated} and Caltech-UCSD Birds-200-2011 (CUB) \cite{WahCUB_200_2011} datasets.
Figure \ref{figure:Flower_CUB} displays additional qualitative results from the Flowers \cite{nilsback2008automated} and Caltech-UCSD Birds-200-2011 (CUB) \cite{WahCUB_200_2011} datasets.\\
\subsubsection{Evaluation Metrics}
\textbf{Intersection of Union (IoU).} %The IoU score measures the overlap of two regions A and B by calculating the ratio of intersection over union, according to
The IoU score quantifies the overlap between two regions. This is achieved by evaluating the ratio of their intersection to their union.
\begin{align}
    \textup{IoU}(A, B) = \frac{|A\cap B|}{|A\cup B|}. \nonumber
\end{align}
%where we use the inferred mask and ground-truth mask as $A$ and $B$ respectively for evaluation.\\
$A$ denotes the ground-truth mask, while $B$ denotes the inferred mask.\\
%as $B$ for assessment purposes.\\
\textbf{DICE score.} Similarly, the DICE score is defined as:
\begin{align}
    \textup{Dice}(A, B) = \frac{2|A\cap B|}{|A|+ |B|}.\nonumber
\end{align}
\noindent
Higher is better for both scores.\\
\subsubsection{Dataset Details}
\textbf{Flowers.} The Flowers \cite{nilsback2008automated} dataset consists of $8,189$ images across $102$ different flower classes. Additionally, it includes segmentation masks generated by an automated algorithm designed explicitly for color photograph flower segmentation \cite{nilsback2007delving}. 
%The images in this dataset have large scale, pose and light variations.\\
The dataset contains images that exhibit substantial variations in scale, pose, and lighting.
Flowers \cite{nilsback2008automated} contains $1,020$ test images.\\
\textbf{CUB.} The CUB \cite{WahCUB_200_2011} dataset contains $11,788$ images covering $200$ bird classes, along with their segmentation masks. 
%Each image is further annotated with $15$ part locations and $1$ bounding box. We use theprovided bounding box to extract a center square from the image, and scale it to $128\times 128$ pixels.
Every image comes with annotations for $15$ part locations, $312$ binary attributes, and $1$ bounding box. We utilized the given bounding box to crop a central square from the image. The CUB dataset includes $1,000$ test images.\\
\textbf{ISIC-2016.} The ISIC-2016 \cite{gutman2016skin} dataset is a public challenge dataset dedicated to Skin Lesion Analysis for Melanoma Detection. Derived from the extensive International Skin Imaging Collaboration (ISIC) archive, it represents a significant collection of meticulously curated dermoscopic images of skin lesions. Within this challenge, a subset of $900$ images is designated as training data, while $379$ images serve as testing data, aiming to provide representative samples for analysis.
%The ISIC-2016 \cite{gutman2016skin} dataset is a public challenge dataset of Skin Lesion Analysis Towards Melanoma Detection released with ISBI 2016. This dataset is based on the International Skin Imaging Collaboration (ISIC) Archive, which is the largest publicly available collection of quality controlled dermoscopic images of skin lesions. The challenge employs a subset of representative images with $900$ images as training data and $379$ images as testing data.

%For all experiments, we resized the input images into a resolution of $256\times 256$ and  generated a $32\times 32$ binary mask for each image utilizing the pre-trained SC-VAE$^\curlywedge$ on ImageNet dataset, a spectral clustering algorithm and boundary connectivity information. The inferred binary mask and ground truth mask were resized to $128\times 128$ to calculate the IoU and DICE scores.
For our experiments, we resized the input images into a resolution of $256\times 256$.
Subsequently, we generated a binary mask of size $32\times 32$ per image by employing the pre-trained SC-VAE$^\curlywedge$ on the ImageNet dataset, along with a spectral clustering algorithm and boundary connectivity information \cite{zhu2014saliency}. To compute the IoU and DICE scores, both the inferred binary mask and the ground truth mask were resized to $128\times 128$.
%\subsubsection{Baseline Methods}
\label{section3}

\clearpage
\clearpage
{
   \small
   \bibliographystyle{ieee_fullname}
   \bibliography{egpaper_arxiv_V2}
}

\end{document}